\documentclass[journal]{IEEEtran}
\usepackage{times}
\usepackage{epsfig}
\usepackage{graphicx}
\usepackage{amsmath}
\usepackage{amssymb}
\usepackage{caption}
\usepackage{subcaption}
\usepackage[]{algorithm}
\usepackage[]{algorithmic}
\usepackage{booktabs}
\usepackage{tabularx}
\usepackage{adjustbox}
\usepackage{accents}
\usepackage{color}
\usepackage{amsmath}
\usepackage{epstopdf}
\usepackage{multirow}
\usepackage{url}
\usepackage{bbm}
\usepackage{flushend}
\usepackage[table]{xcolor}
\newcommand{\first}[1]{\textcolor{red}{#1}}
\newcommand{\second}[1]{\textcolor{blue}{#1}}
\newcommand{\third}[1]{\textcolor{green}{#1}}
\usepackage{multicol}
\usepackage{pbox}

\begin{document}

\title{Good Features to Correlate for Visual Tracking}

\author{Erhan Gundogdu and
A. Ayd\i n Alatan,~\IEEEmembership{Senior Member,~IEEE}


\thanks{E. Gundogdu is with the ASELSAN Research Center, Intelligent Data Analytics
Research Program Department, 06370 Ankara, Turkey, and also with the
Department of Electrical and Electronics Engineering, Middle East Technical
University, 06800 Ankara, Turkey (e-mail: egundogdu87@gmail.com).} 
\thanks{A. A. Alatan is with the Department of Electrical and Electronics
Engineering, Middle East Technical University, 06800 Ankara, Turkey, and
also with the Center for Image Analysis (OGAM), Middle East Technical University,
06800 Ankara, Turkey (e-mail: alatan@metu.edu.tr).}
}

\markboth{IEEE Transactions on Image Processing}{Gundogdu \MakeLowercase{\textit{et al.}}: Good Features to Correlate for Visual Tracking}%

\maketitle

\begin{abstract}
During the recent years, correlation filters have shown dominant and spectacular results for visual object tracking. The types of the features that are employed in these family of trackers significantly affect the performance of visual tracking. The ultimate goal is to utilize robust features invariant to any kind of appearance change of the object, while predicting the object location as properly as in the case of no appearance change. As the deep learning based methods have emerged, the study of learning features for specific tasks has accelerated. For instance, discriminative visual tracking methods based on deep architectures have been studied with promising performance. Nevertheless, correlation filter based (CFB) trackers confine themselves to use the pre-trained networks which are trained for object classification problem. To this end, in this manuscript the problem of learning deep fully convolutional features for the CFB visual tracking is formulated. In order to learn the proposed model, a novel and efficient backpropagation algorithm is presented based on the loss function of the network. The proposed learning framework enables the network model to be flexible for a custom design. Moreover, it alleviates the dependency on the network trained for classification. Extensive performance analysis shows the efficacy of the proposed custom design in the CFB tracking framework. By fine-tuning the convolutional parts of a state-of-the-art network and integrating this model to a CFB tracker, which is the top performing one of VOT2016, 18\% increase is achieved in terms of expected average overlap, and tracking failures are decreased by 25\%, while maintaining the superiority over the state-of-the-art methods in OTB-2013 and OTB-2015 tracking datasets.
\end{abstract}

\let\svthefootnote\thefootnote
\let\thefootnote\relax\footnotetext{1057-7149 ${{\textcopyright}}$ 2018 IEEE. Personal use is permitted, but republication/redistribution requires IEEE permission. See http://www.ieee.org/publications\_standards/publications/rights/index.html for more information. Citation of this paper: E. Gundogdu and A. A. Alatan, "Good Features to Correlate for Visual Tracking," in IEEE Transactions on Image Processing, vol. 27, no. 5, pp. 2526-2540, May 2018. doi: 10.1109/TIP.2018.2806280,
URL: http://ieeexplore.ieee.org/document/8291524}

\addtocounter{footnote}{0}\let\thefootnote\svthefootnote


\IEEEpeerreviewmaketitle
\section{Introduction}
\label{secIntro}
One of the major problems in computer vision is single object visual tracking, which has potential applications including visual surveillance, security and defense applications and human computer interaction. Although the definition of this problem varies according to the application and the type of the target object, it can be described as tracking an object, which is marked by the user at the beginning of a video sequence. Tracking is accomplished by predicting the state of the object at each frame. The benchmark datasets  \cite{Benchmark2013} and \cite{BenchmarkPAMI}, which are useful tools to assess the performances of the tracking algorithms, define the ground truth object state as the bounding box surrounding the object in the image domain. Thus, if there is more overlap between the prediction and the ground truth bounding box, more accurate localization of the target is obtained. In order to improve the accuracy of the tracking, various machine learning concepts have been borrowed, such as sparse generative methods \cite{L1APG, L1Other}, support vector machines \cite{STRUCK} and deep learning \cite{MDNET, PorikliBMVC2}.
\begin{figure}[ht]
\centering
\includegraphics[width=0.48\textwidth]{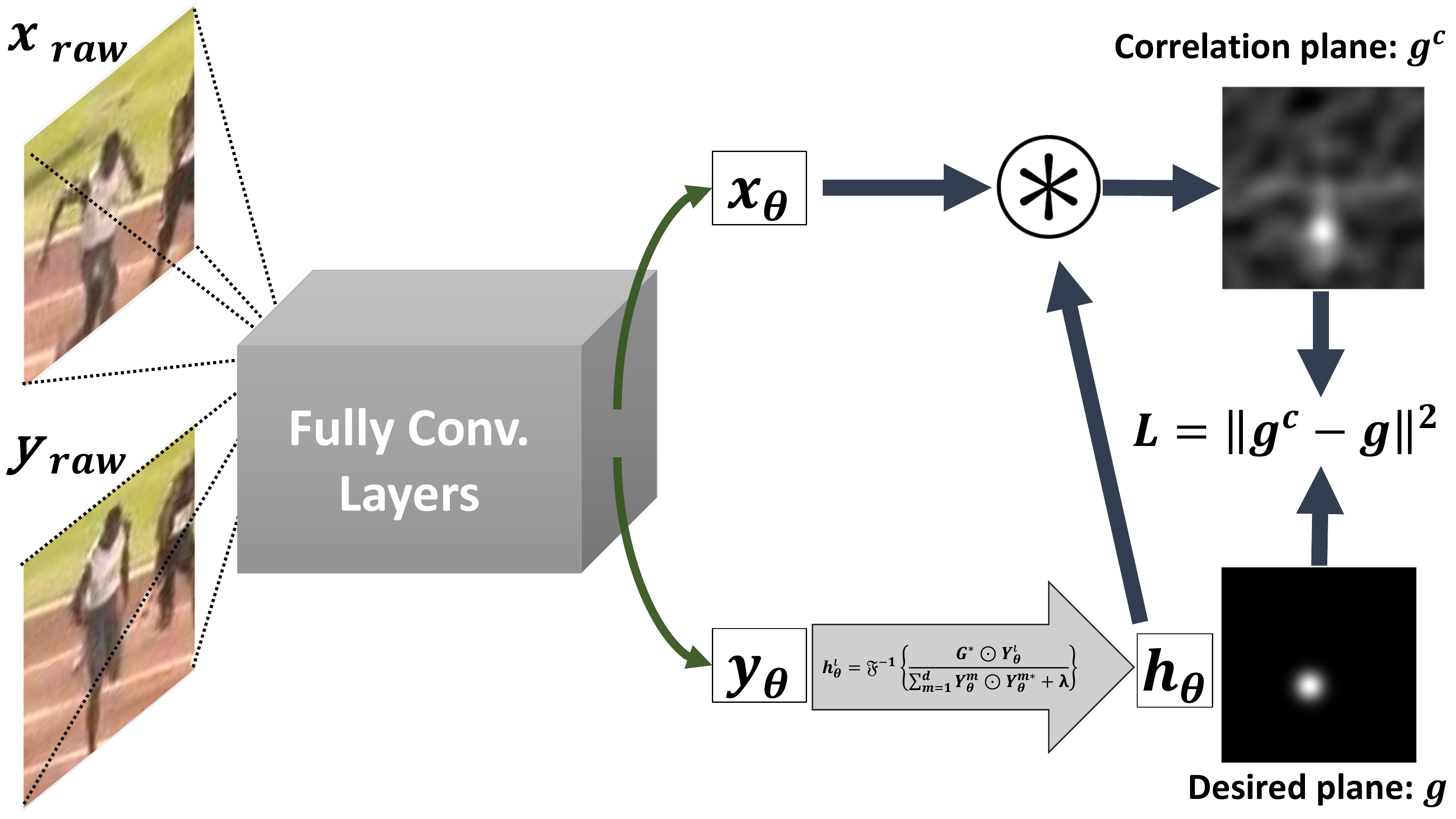}
\caption{\small Method overview: A CNN model is trained to improve the correlation quality such that when the correlation filter $h_\theta$ and the observation $x_\theta$ (both depend on the parameters of the CNN model) are circularly correlated, the resulting correlation output will be improved. This can be achieved by training the model with appropriate set of $x_{raw}$ and $y_{raw}$ pairs which differ in the appearance and translation of the object with respect to the center of the patch.}
\label{algflow}
\normalsize
\end{figure}

During the last decade, a substantial amount of effort has been put on the correlation filter based (CFB) trackers, while the pioneering study of Bolme \emph{et.al.} \cite{MOSSE} has triggered the use of correlation filters for visual tracking. Concretely, the attractive participants and the winners of the Visual Object Tracking (VOT) challenges are from the CFB tracking family in the last three years. This being the case, various improvements over the base correlation filter formulation are frequently proposed to enhance the accuracy of the tracking. The best performing trackers of VOT2015 \cite{VOT2015} and VOT2016 \cite{VOT2016} challenges utilize the pre-trained deep features \cite{VGG} specifically trained on the large scale image recognition datasets \cite{ImageNet,ILSVRC2015} for object classification. In order to employ these networks to the CFB frameworks, only convolutional layers are utilized, since the shift invariance property is intended to be maintained due to the nature of the correlation operation. The correlation capability of the features are limited to the classification network, which will hopefully generate good features to correlate. Nevertheless, learning deep convolutional features for CFB tracking cost function is still unexplored. 

In order to break the limits of the aforementioned models of object recognition, we address the problem of learning a fully convolutional neural network which generates useful feature maps for correlation operation. The proposed framework consists of a single fully convolutional network. Training of the model is performed by propagating two image patches, which contain the same visual object, through the model. Once the feature maps of the top layer are obtained for each image patch, the correlation filter is calculated from the template patch, and the difference between the estimated and desired correlation response is to be minimized. The reduction of the difference between these two signals is obtained by the backpropagation of the error and the stochastic gradient descent procedure as in the case of training a classification based CNN architecture.

In this study, our contributions are summarized as follows:
\begin{itemize}
\item A framework to train a fully convolutional deep network is presented for the correlation filter cost function.
\item We provide an efficient formulation to backpropagate the network according to the CFB loss.
\item In order to efficiently train networks with large number of feature channels at their final layer, we propose to include an auxiliary layer with few number of feature channels as the final layer of the network.
\item The network trained on the dataset generated for our specific scenario is integrated into the CFB tracking methods DSST \cite{DSST}, SAMF \cite{SAMF} and CCOT \cite{CCOT}. This significantly boosts the performance of the integrated trackers in benchmark datasets, VOT2016, OTB-2013 and OTB-2015\footnote{The source code and the presented results are publicly available. Please check https://github.com/egundogdu/CFCF}.
\end{itemize}



In the remaining part of the manuscript, we first present the closely related work to our visual tracking framework in Section \ref{secRelated}. The ultimate goal is to obtain robust feature maps to be employed in CFB trackers. Hence, the CFB formulation is explained in Section \ref{secCFB}. In Section \ref{secProposed}, the proposed feature learning method is given with detailed derivations. Section \ref{secExperiments} reports the experimental results as well as the implementation details and dataset generation. Finally, Section \ref{secConclusion} discusses conclusive remarks about the proposed methodology and promising feature work.

\section{Related Work}
\label{secRelated}
Numerous methods have been proposed to solve the visual tracking problem for decades. In this section, our aim is to give a literature survey as comprehensive and recent as possible to link the proposed method and the state-of-the-art trackers.
\subsection{Discriminative Trackers}
Discriminative methods utilize a classifier model, which is responsible for the classification of a visual sample as either the object or background. Model training is performed by collecting positive and negative examples from the region of interest that is provided at the beginning of the tracking. The object localization is generally performed by looking for the candidate location with the highest classifier score. The Haar-like features \cite{HAAR} and/or Local Binary Patterns \cite{LBP} are employed in \cite{MIL} and \cite{OAB} to train an ensemble of classifiers, since these features can be efficiently extracted by integral images. Support Vector Machines are employed in \cite{STRUCK} and further improved in \cite{StruckPart,DLSVM}. Deep learning based track-by-classification methods are also studied in the works, such as \cite{PorikliBMVC2} and \cite{MDNET}. Nevertheless, discriminative methods must evaluate their classifiers at each candidate location, thus bringing a significant computational load.
\subsection{Generative Methods}
Unlike the discriminative tracking approaches, generative methods describe appearance model for the object and optionally for the background. The object location is estimated as the one which contains the test instance with the most similarity to the appearance model. The model is updated with the object instance gathered from the predicted location. The study in \cite{IVT} proposes an online subspace learning method. Another method in \cite{KBOT} models the object appearance in terms of the brightness histogram of the object patch. On the other hand, sparse visual trackers are proposed in \cite{L1APG,L1Other,L1} and \cite{L1Other2}, which mainly obtain a sparse projection of the object instances with respect to a dictionary consisting of the object templates. In \cite{SparseMultiTask} and \cite{SparseLowRank}, a joint sparsity constraint is forced in such a way that the resulting sparse coefficients are not only sparse themselves, but also their usage for different samples are sparse as well. Non-negative matrix factorization is also casted to the visual tracking problem in \cite{NMF} to learn a non-negative dictionary. Generative methods suffer from the same problem, the evaluation of the objectness at each candidate location, as the discriminative trackers do.
\subsection{Correlation Filter Based Trackers}
Correlation filters have become popular by the pioneering study in \cite{MOSSE}, which mainly attempts to minimize the sum of squared error between the desired correlation response and the circular correlation of the filter and the object patch. Utilizing the Convolution Theorem and properties of Fast Fourier Transform, the minimization of correlation filter cost is efficiently computed in the frequency domain. The work in \cite{DSST} extends \cite{MOSSE} by formulating the multi-channel support and employing HOG feature maps. In addition, the method in \cite{DSST} has a multi scale search support to estimate the scale of the object and to increase the tracking performance. 

Kernelized correlation filters (KCF) are proposed in \cite{circulant}. \cite{KCF} generalizes \cite{circulant} for multi-channel support. Various extensions of KCF is proposed in \cite{dpKCF}, \cite{sKCF} and \cite{PartBasedCorrelation} for scale estimation as well as part-based proposal combinations. The imperfect training example issue is addressed in \cite{LimitedBoundaries} and \cite{SRDCF} by applying a spatial regularization on the corelation filter to increase the search range. In \cite{ChanSpatial}, spatial constraints are forced on the correlation filter. The study in \cite{TRA} simultaneously learns the correlation filter and the desired correlation response to circumvent the problems of circularly shifted training patches. To distinguish the object from the background, color statistics are used in \cite{STAPLE} as a complementary model to correlation filters. On the other hand, the tracker in \cite{CACF} learns the correlation filter by considering the background patches surrounding the target object, and achieves a good trade-off between computational complexity and accuracy.

Pre-trained deep CNN models are utilized in \cite{HCF} and \cite{deepSRDCF} as the feature maps to correlate. Moreover, the method in \cite{CCOT} presents a continuous domain correlation filter learning to address the utilization of feature maps with different resolution.

Concurrently with our proposed method, two independent studies (\cite{CFNet} and \cite{DCFNet}) have recently been presented to train fully convolutional networks for correlation filter based tracking. Valmadre et al. \cite{CFNet} propose to learn a fully convolutional network along with the element-wise logistic loss function while considering the correlation filter formulation in their framework as we do. However, our proposed cost function reduces the squared error between the desired and estimated correlation responses, which is the multiple channel correlation filter cost function. Moreover, we train a larger network model than those of \cite{CFNet} and \cite{DCFNet}, and observe a significant amount of tracking performance increase in benchmark datasets, while \cite{CFNet} and \cite{DCFNet} mainly concentrate on lightweight architectures for less computational complexity. In this study, the backpropagation formulations are based on the generalized chain rule and the conjugate symmetry property of real signals in the Fourier domain, however, the backpropagation derivations are performed in \cite{CFNet} by exploiting the adjoint of differentials. Furthermore, the proposed framework has been trained on our generated dataset from ILSVRC Video dataset, which attempts to handle the imperfect training example issue, whereas both the test image and the template patch have the target object centered in \cite{CFNet}, and a relatively smaller dataset is adopted in \cite{DCFNet}. Finally, our proposed feature learning framework achieves favorable performance against these concurrent works since the proposed method mainly focuses on training a relatively larger network, and achieves the state-of-the-art results in benchmark tracking datasets.

\subsection{Custom Architectures for Visual Tracking}
Recently, various deep architectures with customized layers or objective functions have been proposed. An application of Siamese feature learning to the visual object tracking is proposed in \cite{SINT} where the network learns to output similar features for various appearances of the target object and dissimilar ones for the target and non-target samples. Nevertheless, evaluation of many candidates are quite expensive. Hence, a CNN model is introduced in \cite{GoTurn}, which directly learns to output the relative location of the object with respect to a reference object instance and avoids the expensive candidate evaluations and the feature matching phase. Unlike the model in \cite{GoTurn} employing fully connected layers, a fully convolutional neural network is presented in \cite{SiamFC}. In this approach \cite{SiamFC}, the object template and the test frame are passed through the same convolutional layers, and their correlation is obtained by the sliding window approach. The sliding window stage is operated in the convolutional layer format, since the standard deep learning libraries are efficiently exploited in order not to sacrifice much from the computation time, while still suffering from the satisfactory tracking performance.

Another popular concept is Recurrent Neural Networks (RNNs) \cite{RNNBook}, which is a useful neural network model, especially in natural language processing. RNNs are employed in \cite{RTT} in order to estimate the confidence map of the target object by modeling the spatial relationships between the object and the background. Another spatial perspective is to spatially model the object structure \cite{SANet}. This study successfully applies this idea to the visual tracking problem in order to assist the CNN layers. Unlike the use of RNNs for the spatial relationships, \cite{RATM} and \cite{RATMlike} propose to learn an RNN model to directly estimate the motion of the object by modeling their RNNs to learn the relationships between the frames sequentially. Nevertheless, the visual tracking experiments are conducted on the simulation data, and they lack the performance on the standard benchmarks, such as VOT challenges \cite{VOT2014,VOT2015,VOT2016} or Online Tracking Benchmarks \cite{Benchmark2013,BenchmarkPAMI}.

\subsection{Combining Trackers}
Combination of multiple online trackers is another research path. For instance, multiple correlation trackers are run at different parts of the object in \cite{PartBasedCorrelation}. A part-based version of MOSSE \cite{MOSSE} has been proposed in \cite{POSSE} to accomplish object detection task. Reliable patches are tracked in \cite{RPT} using KCF \cite{KCF} as the base tracker. The work in MEEM \cite{MEEM} selects the best SVM-based discriminative tracker according to an entropy minimization criterion. Markov Chain Monte Carlo sampling is also used to sample trackers and combine them \cite{expert2}. On the other hand, various trackers with mixed feature types are combined in \cite{expert3}. Hybrid methods combining generative and discriminative approaches are proposed in \cite{expert5, expert4}.
Since deep discriminative networks \cite{MDNET} have an impact in the visual tracking literature, in \cite{TCNN}, a tree-structure stores different appearances in the nodes of the tree as CNN models. This provides a robustness to significant appearance changes, while suffering from the heavy computational load.

\section{Correlation Filter Formulation}
\label{secCFB}
In this section, we briefly summarize the two correlation filter based tracking methods, Discriminative Scale Space Tracker (DSST) \cite{DSST} and Continuous Convolution Operator Tracker (CCOT) \cite{CCOT} for completeness. The learned features from the proposed framework are integrated into these trackers due to their notable performance in the benchmark sequences.
\subsection{Multiple Channel Linear Correlation Filters}
\label{secDSST}
DSST \cite{DSST} is the multiple channel extension of MOSSE \cite{MOSSE}. The feature maps $\{y^1, ...,y^d\}$ correspond to the training example $y$, which consists of particular feature maps, such as HOG orientation maps or deep feature maps with the same dimension as the object patch. The desired correlation mask of the training example $y$ is denoted by $\hat{g}$.
\begin{equation}
	\label{eqDSST}
	\mathcal{L}(h_t) =  {\sum\limits_{i=1}^t\left\lVert(\sum\limits_{l=1}^d h_t^l\circledast y_i^l)-\hat{g_i}\right\rVert}^2 + \lambda_\epsilon{\sum\limits_{l=1}^d \lVert h_t^l\rVert}^2
\end{equation}
Here, $\lambda_\epsilon$ is the control parameter for $l_2$ regularization term of the filter, $\circledast$ is the circular correlation operation, which can be described as:
\begin{equation}
\label{eqCorrTheorem}
a \circledast b = \sum\limits_{i} a[i]b[n+i]=\mathcal{F}^{-1}\{A^* \odot B\},
\end{equation}
where ${F}^{-1}$, $\odot$ and $^*$ are the inverse DFT, element-wise multiplication and the conjugation operations, respectively. The lowercase letters represent the signals in the spatial domain, whereas the uppercase letters denote the signals in the DFT domain.

Moreover, the subscripts of the variables indicate the time indices, \emph{i.e.}, $h_t^l$ is the $l^{th}$ feature channel of the corelation filter calculated at time $t$. As \eqref{eqDSST} suggests, a set of filters $\{h_t^l\}_{l=1}^d$ are to be estimated such that the correlation operation between $h_t^l$'s and $x_i^l$'s are summed and the error between the desired response $\hat{g_i}$'s and the summed correlation results $\sum\limits_{l=1}^d h_t^l\circledast x_i^l$ should be minimized under the $l_2$ regularization of the correlation filters. There exists a closed form solution in the frequency domain for one training example, \emph{i.e.} $t=1$:
\begin{equation}
\label{eqDSSTFinal}
H^l = \frac{Y^l\odot \hat{G^*}}{\sum\limits_{k=1}^d Y^k\odot Y^{k*}+ \lambda_\epsilon}, \forall l \in {1,..., d},
\end{equation}
where the time dependency is dropped for convenience. At time $t$, the filter $H_t^l$ is updated by applying moving average to the numerator and denominator of \eqref{eqDSSTFinal} separately via the following relations:
\begin{equation}
\begin{split}
\label{eqDSSTUpdate}
A_t^l = (1-\mu)A_{t-1}^l+\mu \hat{G_t^{*}}\odot Y_t^l,\\
B_t = (1-\mu)B_{t-1}+\mu \sum\limits_{k=1}^d Y_{t}^k\odot Y_{t}^{k*},
\end{split}
\end{equation}
where $\mu$ is the model update rate. The correlation of an object patch $z$ and the model $H_t^l$ is calculated by using the updated numerator $A_t^l$ and denominator $B_t$ of $H_t^l$ in the frequency domain using:
\begin{equation}
\label{eqDSSTLoc}
c=\mathcal{F}^{-1}\{({\sum\limits_{l=1}^d A^{l*}_t\odot Z^l})/({B_t+\lambda_\epsilon})\},
\end{equation}
where the spatial domain correlation mask is obtained by taking the inverse Fourier transform.
The new location of the object in the next frame is estimated as the location giving the maximum value at $c$ in \eqref{eqDSSTLoc}. 

For scale estimation, DSST extracts $\tilde{d}$-dimensional HOG features for $S$ scale factors. The base target size is multiplied by the scale factor. The corresponding region is cropped and described by $\tilde{d}$-dimensional features similar to the location estimation procedure. Then, the scale correlation filter $h_{scale}$ is calculated for the scale samples $y_s \in \mathcal{R}^{\tilde{d}\times S}$. The optimal scale is determined as the scale index giving the highest value on the correlation response of the test instance $z_{scale}$ and $h_{scale}$.

\subsection{Continuous Convolution Operators for Visual Tracking}
\label{secCCOT}
A continuous domain formulation for correlation filters is proposed in CCOT \cite{CCOT} to combine feature maps of different resolutions, especially deep feature maps at different layers.

Unlike the constant dimension assumption for all of the feature maps, each training sample $y_j$ is allowed to have the feature maps with different dimensions as $y_j^{d} \in \mathcal{R}^{N_d}$. To implicitly model the signals in the continuous domain, the interval $[0,T)$ is assumed to be the support interval. For each feature map $d$, the interpolation operator is expressed as:
\begin{equation}
\label{eqCCOTInterp}
J_d\{y^d\}(t) = \sum\limits_{n=0}^{N_d-1}y^d[n]b_d \left(t-\frac{T}{N_d} n \right),
\end{equation}
where $b_d \in L^2(T)$ is the interpolation function. A linear convolution (or a correlation) operator $S_f$ is required such that a sample $y$ is mapped to a target confidence response $s(t)=S_f\{y\}(t)$. Since there exist $d$ feature maps, the correlation filters $f = (f^1, f^2, ... f^D) \in L^2(T)^D$ is intended to be estimated. The convolution operator in the continuous domain is described as:
\begin{equation}
\label{eqCCOTConvOp}
S_f\{x\}=\sum\limits_{d=1}^D f^d \ast J_d\{y^d\}
\end{equation}
In the above relation, $\ast$ is the continuous domain correlation. Although the initial signals are discrete, they are first converted to the continuous domain by using the operation $J_d\{y^d\}$. Moreover, there should be continuous desired values $g_j$ for each training example $y_j$. The correlation filter cost function is defined in the continuous domain by:
\begin{equation}
\label{eqCCOTCost}
E(f) =  \sum\limits_{j=1}^m \alpha_j \lVert S_f\{y_j\}-g_j \rVert^2 + \sum\limits_{d=1}^D \lVert w f^d \rVert^2,
\end{equation}
Here, $\alpha_j$ represents the importance of the sample $y_j$, and $w$ is a spatial penalty function to regularize the correlation filter in the spatial domain for suppressing the boundaries.

In order to learn the filter $f$ minimizing the cost in \eqref{eqCCOTCost}, the operations are projected to the discrete frequency domain. Then, the cost in \eqref{eqCCOTCost} is converted to a set of normal equations. The Conjugate Gradient Descent is utilized to iteratively optimize this cost. The implementation details can be found in \cite{CCOT}. Once the object is localized, a multi-scale search is adopted with $S$ scales to find the best matching scale by looking at the correlation response at every scale.

Thus far, the correlation filter based tracking methods that are tested in this study are summarized and these techniques are utilized to assess the effectiveness of the proposed feature learning method. The proposed framework is presented next.
\section{Proposed Framework for Feature Learning}
\label{secProposed}
\subsection{Preliminaries}
In order to perform the training of the proposed framework in Figure \ref{algflow}, a set of triplet training samples is required. A triplet is represented by $\mathcal{T}_i \triangleq \{x_i,y_i,g_i\}$. $y_i$ is the template image patch which contains the object at its center. $x_i$ is the test image patch including the non-centered object. $g_i$ is the desired correlation response which has a peak at the location shifted from the center of the patch by the amount of the correct motion of the object between $x_i$ and $y_i$. Throughout the intermediate derivation and equivalences between them, these three discrete signals are assumed to be $1$-dimensional. The derivations are also valid for $2$-dimensional case, since all of the utilized operations are separable for horizontal and vertical dimensions, such as Discrete Fourier Transform (DFT). For a signal $x$, $x[n]$ denotes its $n^{th}$ component, and $x[n+i]$ is its shifted version by an integer amount $i$ to the left circularly. The circular shift is important to exploit the Correlation Theorem for real signals.

It is notable that a feature generation function $f(.)$ of the image patch $I$, which is typically integrated into the CFB trackers, should carry the shift invariance property, i.e. if $I_\theta[x][y]=f(I[x][y])$ and $Y_\theta[x][y]=f(I[x-k\delta_x][y-k\delta_y])$, then $Y_\theta[x][y]\approx I_\theta[x-\delta_x][y-\delta_y]$ should be satisfied, where $I[.][.]$ is a 2-D discrete signal and $k$ is the scale factor of the transformation function $f(.)$. Thus, we employ fully convolutional CNN models, which contain convolutional, batch normalization, pooling and ReLU layers. These layers do not violate this property.
\subsection{The Proposed Loss Function for Parameter Learning}
The proposed learning methodology utilizes the stochastic gradient descent (SGD) as in most of the deep learning frameworks. Our cost function for $N$ triplet examples is defined as:
\begin{equation}
\label{eqCost}
\mathcal{L} = \sum\limits_{i=1}^N \mathcal{L}(\theta)_i,
\end{equation}
where
\begin{equation}
\label{eqCosti}
\mathcal{L}(\theta)_i={\left \| \sum\limits_{l=1}^d h^l_i (\theta) \circledast x^l_i(\theta) - g_i\right \|}^2
\end{equation}
In the equation above, $x_i^l(\theta)$ is the network output with parameters $\theta$ for the input patch $x_i$. In \eqref{eqCosti} and \eqref{eqCost}, $\theta$ represents the parameters of the fully convolutional network model. As it is given in \eqref{eqDSSTFinal}, $h^l_i (\theta)=\mathcal{F}^{-1}\{{(Y^l_i(\theta)\odot \hat{G_i^*})}/({\sum\limits_{k=1}^d Y^k_i(\theta)\odot Y_i^{k*}(\theta)+ \lambda_\epsilon})\}$ is the minimizing correlation filter for the feature map $l$, which is a function of $\{y_i^l(\theta)\}_{l=1}^d$ ($y_i^l(\theta)$ is the output of the network for $y_i$). Note that, in this formulation, $\hat{G_i}$ represents the DFT of the centered desired Guassian shaped correlation response for the template $y_i$ as in \eqref{eqDSSTFinal}. The goal of the proposed method is to learn appropriate values for $\theta$ that will help to reduce the cost in \eqref{eqCost}.

The major difference between the correlation filter cost in \eqref{eqDSST} and the proposed one in \eqref{eqCost} is that the cost in \eqref{eqCost} is minimized with respect to the network parameters $\theta$ for the given correlation filter solution in \eqref{eqDSSTFinal}, whereas the cost in \eqref{eqDSST} is minimized with respect to the correlation filters $\{h^l_i\}_{l=1}^d(\theta), \forall i$. The regularization part in the second term of \eqref{eqDSST} is removed in the proposed cost function, since the correlation filter solution in \eqref{eqDSSTFinal} already penalizes the norm of the correlation filters.
We hypothesize that the correlation quality will increase during test time in a visual tracking application, if the proposed cost function is reduced with respect to the parameters of the network by a stochastic training process on an appropriately generated dataset.
\subsection{Gradient of the Loss, $\mathcal{L}(\theta)_i$}
In order to learn a model with parameter set $\theta$, the gradient of the loss with respect to $\theta$ is required. By the multivariable chain rule, the gradient of the loss in \eqref{eqCosti} can be written as\footnote{For the convenience of the derivations, $\theta$ and the subscript $i$ representing the index of the training example are dropped from the variables.}:
\begin{equation}
\label{eqGradCosti}
\nabla_\theta\mathcal{L}=\sum\limits_l\frac{d\mathcal{L}}{d x^l}\frac{d x^l}{d \theta} + \sum\limits_l\frac{d\mathcal{L}}{d y^l} \frac{d y^l}{d \theta},
\end{equation}
By applying the multivariable chain rule again, the first multiplicand in the second term of \eqref{eqGradCosti} becomes:
\begin{equation}
\label{eqGradSub}
\frac{d \mathcal{L}}{d y^l} = \sum\limits_{k=1}^d \frac{d \mathcal{L}}{d h^k} \frac{d {h^k}}{d y^l}
\end{equation}
If the error signal $e[n]$ is defined as:
\begin{equation}
\label{eqApp4}
e[n]=\sum\limits_l \sum\limits_i h^l[i]x^l[i+n]-g[n],
\end{equation}
the terms $\frac{d\mathcal{L}}{dx^l}$ and $\frac{d\mathcal{L}}{dh^k}$ in \eqref{eqGradCosti} and \eqref{eqGradSub} can be written as:
\begin{equation}
\label{eqFromAppendix}
\begin{split}
\frac{d \mathcal{L}}{d h^k}=\mathcal{F}^{-1}\{E^* \odot X^k\}
\\
\frac{d \mathcal{L}}{d x^l}=\mathcal{F}^{-1}\{E \odot H^l\}
\end{split}
\end{equation}
The partial derivatives in \eqref{eqFromAppendix} are derived in Appendix \ref{secAppendix1}. The Jacobians of the vectors $y^l$ and $x^l$ with respect to the model parameters $\theta$ ($\frac{d y^l}{d \theta}$ and $\frac{d x^l}{d \theta}$) can be efficiently calculated by using the standard backpropagation tools of the existing deep learning libraries. 

Till now, all of the terms to calculate the gradient in \eqref{eqGradCosti} are presented, except for the Jacobian $\frac{dh^k}{dy^l}$. Since the relations between $h^k$ and $y^l$ are in the DFT domain as described in \eqref{eqDSSTFinal}, this term should be first converted to the DFT domain as follows:
\begin{equation}
\label{eqJacobDFT}
\frac{dh^k}{dy^l} = \frac{dh^k}{dH^k} \frac{dH^k}{dY^l} \frac{dY^l}{dy^l} = F^H \frac{dH^k}{dY^l} F,
\end{equation}
where $F$ and $F^H$ are DFT and inverse DFT matrices, respectively. The relation between $H^k$ and $Y^l$ are expressed independently for each frequency component as in \eqref{eqDSSTFinal}. Hence, the derivative of the division rule enables us to write:
\footnotesize
\begin{equation}
\label{eqKYJac}
\begin{split}
\frac{d H^k}{d Y^l}= \mathcal{I}(k==l) diag\left(\hat{G^*}/\left(D(Y)\right)\right) \\ -diag\left(\frac{\hat{G^*} \odot Y^k \odot {Y^l}^*}{(D(Y))^2}\right)\\ -diag\left(\frac{\hat{G^*} \odot Y^k \odot Y^l}{(D(Y))^2}\right) M
\end{split}
\end{equation}
\normalsize
where $\mathcal{I}(.)$ is the indicator function yielding 1 when its argument is true, and 0 otherwise, and $D(Y)={\sum\limits_{m=1}^d Y^m \odot {Y^m}^*+\lambda}$. In the above relations, the signals are treated as individual complex variables, and $M$ is the matrix for the circular time reversal operation, equal to $\frac{d Y^{l*}}{d Y^l}$ (\emph{i.e.} the Jacobian of $Y^{l*}$ with respect to $Y^l$) due to the conjugate symmetry property of real signals in DFT domain. In other words, $Mv$ is the time reversed version of the signal $v$ by fixing its first element.

If the following intermediate signals are defined:
\begin{equation}
\label{eqIntermed}
\begin{split}
K_1^{kl}=\mathcal{I}(k==l) \frac{\hat{G^*}}{D(Y)}\\ K_2^{kl}=\frac{\hat{G^*} \odot Y^k \odot {Y^l}^*}{(D(Y))^2} \\ K_3^{kl}=\frac{\hat{G^*} \odot Y^k \odot Y^l}{(D(Y))^2},
\end{split}
\end{equation}
then \eqref{eqJacobDFT} can be simplified to:
\begin{equation}
\label{eqKYJacSimplified}
\frac{d h^k}{d y^l} = F^H(diag(K_1^{kl}-K_2^{kl})-diag(K_3^{kl})M)F
\end{equation}

All of the operations performed in the DFT domain are element-wise multiplication of the signals or their reciprocals, conjugation operation and so on, which do not violate the real property of the resulting signals. Hence, the conjugation operation in the spatial domain keeps the imaginary parts of the gradient terms to be zero.

Finally, if $\frac{d h^k}{d y^l}$ in \eqref{eqKYJacSimplified} and $\frac{d\mathcal{L}}{dh^k}$ in \eqref{eqFromAppendix} are replaced in \eqref{eqGradSub}, and the Hermitian operation is taken for the overall expression,
\begin{equation}
\label{eqGradSubF}
\nabla_{y^l}\mathcal{L}=\frac{d \mathcal{L}}{d y^l} = \mathcal{F}^{-1}\{\sum\limits_{k=1}^d (K_1^{kl}-K_2^{kl})^* \odot A^k-(K_3^{kl}\odot {A^k}^*) \}
\end{equation}
is obtained as the gradient of the loss in \eqref{eqCosti} for the $l^{th}$ feature map of the template image patch $y$. In \eqref{eqGradSubF}, $A^k$ stands for the DFT of $\frac{d \mathcal{L}}{d h^k}$.

During the training process, the gradient of the cost with respect to the parameters and the activations of the network are computed as explicitly derived and formulated above for $B_N$ triplet examples ($\{\mathcal{T}_i\}_{i=1}^{B_N}$) in a batch. Then, the GD optimization is performed for all of the randomly sampled batches. 

It should be noted that the proposed method can be modified so that feature channels are independent from each other. However, treating the feature channels independently makes it difficult to reduce the loss function due to the decrease of the frequency components in the denominator of \eqref{eqDSSTFinal} and probably causing the gradient overflow in spite of the careful selection of learning rate. Moreover, the employed formulation is optimal according to \eqref{eqDSST}. Due to the aforementioned reasons, we prefer the current loss and leave the independence assumption on the feature channels as a future study.
\subsection{Computational Complexity and Its Reduction}
\label{secComplexity}
It is notable that all of the necessary gradient terms in \eqref{eqGradCosti} and \eqref{eqGradSub} can be efficiently computed in the DFT domain with only element-wise multiplications, divisions, summations and DFT transform operations. The main computational burden results from the DFT calculation with the complexity of $\mathcal{O}({Plog(P)})$ ($P$ is the length of the signal). Moreover, the operation in \eqref{eqGradSubF} is performed for each feature index out of $d$ feature maps. Hence, the final complexity of backpropagating one triplet through the network has the complexity of $\mathcal{O}(d{Plog(P)})$. Depending on the value of $d$ (typically ranging between $64$ and $512$), this complexity could be impractical. In our Matlab implementation, the training speeds are $36$, $16$ and $6$ frames per second for $d$ values $32$, $64$ and $128$, respectively.

To train the convolutional parts of VGG as fast as the classification networks such as Alexnet \cite{ImageNetCNN} and VGG \cite{VGG}, an auxiliary layer with relatively fewer feature maps has been added on top of the $conv-5$ layer to reduce the computation time. It is observed that the robustness of the localization improves as the number of feature maps increase \cite{CCOT,deepSRDCF}. However, it can be claimed that if the correlation quality of a layer is enhanced, then this quality is expected to be transferred to the lower layers. This claim is analyzed in Appendix \ref{Appendix2} for a layer with two feature maps and a layer with single feature map which is the summation of the two feature maps of the previous layer under mild assumptions. Moreover, the amount of quality improvement reduces as the distance between the layers increases. In this way, the outputs of all the convolutional layers before top one can be used as ``good'' features without increasing the complexity of the training stage.

In the following section, the implementation details and the conducted experiments are presented to show the validity of our approach.
\section{Experimental Results}
\label{secExperiments}
\subsection{Performance Evaluation}
The proposed tracker configurations are evaluated on OTB-2013 \cite{Benchmark2013}, OTB-2015 \cite{BenchmarkPAMI} and VOT2016 \cite{VOT2016} datasets. OTB-2013 is a subset of OTB-2015 whereas VOT2016 is the 2016 challenge dataset of the Visual Object Tracking (VOT) committee.

For OTB, there exist two main performance metrics. 1) Success curve is computed by the ratio of successfully tracked frames according to a threshold on the overlap ratio, which is defined as the intersection over union of the predicted and ground-truth bounding boxes. The trackers are ranked according to the Area-Under-Curve (AUC) score of the success curve. Overlap precision (OP) is also a respectable metric which orders the trackers according to the value of the average success on the threshold $0.5$. 2) Precision curve is plotted according to the center localization error and the ratio of the frames with a localization error below a threshold is accept as the distance precision (DP). The curve is plotted by varying the threshold and the trackers are ranked according to the average distance precision value at $20$ pixels.

VOT2016 has a different tracking assessment technique including three major metrics. 1) Accuracy is the mean intersection over union of the frames in a sequence. 2) Failure is the mean number of failures per sequence. These two metrics are raw metrics. The ranking of a particular metric (failure or accuracy) is obtained by ordering the compared trackers with respect to that metric, and the statistically significant tracker rankings are merged. 3) Expected average overlap (EAO) is estimated for a selected range of sequence lengths. Concretely, a specific expected average overlap $\phi_{N_s}$ is estimated by averaging the accuracy values in the segments that are longer than $N_s$ while discarding the segments shorter than $N_s$ with no failure termination. The segments shorter than $N_s$ with a failure are zero-padded; hence, penalizing the failure case for that particular $N_s$ length. These $\phi_{N_s}$ values are determined for the set $\{\phi_{N_s}\}_{N_{lo}}^{N_{hi}}$ and the final score is the mean of these expected values in the set. $N_{lo}$ and $N_{hi}$ are determined according to the sequence length histogram of the dataset.

\subsection{Dataset Generation}
The proposed method is realized by generating two datasets with appropriate fully convolutional models. For the first dataset,
we generate 200K training examples by utilizing the VOT2015 dataset \cite{VOT2015}, consisting of 60 sequences with different attributes. The bounding box of each object is provided for each frame. We crop approximately two times larger area of the object size and resize the images to the appropriate size of the network (101$\times$101 in our experiments). To keep the aspect ratio of the objects, we crop the squares from the region of interests of the object, where the side length of the square is $2\sqrt{W H}$ ($W$ and $H$ are the width and height of the object, respectively.). Generated $y_{raw}$ centers the object, since these patches are indeed templates for us. However, $x_{raw}$ is obtained by shifting the center of the object, since our aim is to break the influence of the circular translation over the actual translation. The shift amount is determined by a random variable which is uniformly distributed in the range of values $[-0.3\times W, 0.3\times W]$ and $[-0.3\times H, 0.3\times H]$ for horizontal and vertical translations. The frame difference between $y_{raw}$ and $x_{raw}$ is a Gaussian random variable with standard deviation of $5$ frames. We entitle this dataset as Convolutional Features for Correlation Filters VOT2015 (\textbf{CFCF VOT2015} for short). 

The custom architecture model has been trained on a dataset generated by using VOT2015 dataset including $60$ video sequences. $11$ of the sequences in VOT2015 also exist in OTB-2013 Benchmark dataset \cite{Benchmark2013}. Thus, it prevents the evaluation to be fulfilled on the full OTB-2013 sequences. Moreover, VOT2015 is not a large-scale dataset even though the generated samples are over $200K$. This situation discourages to train or fine-tune the state-of-the-art convolutional networks such as \cite{VGG, VGG_verydeep}. In order to handle this situation, a new dataset is generated from the large-scale video sequences of ILSVRC challenge dataset \cite{ILSVRC2015}. 

The existing benchmarks OTB-2013\cite{Benchmark2013}, OTB-2015 \cite{BenchmarkPAMI}, VOT2014 \cite{VOT2014}, VOT2015 \cite{VOT2015}, VOT2016 \cite{VOT2016}, NUS-PRO \cite{NUS-PRO} and ALOV \cite{ALOV} have limited number of sequences (less than a thousand). Moreover, some of these datasets have overlapping sequences (\emph{e.g.}, VOT2015 and VOT2016 is a subset of ALOV and OTB datasets.). Thus, the total number of  sequences are not large enough to conveniently train a large network model. As a secondary note, all of these datasets contain sequences with similar appearances and challenges, hence increasing the risk of memorization. Due to the aforementioned reasons, in our study, we also employ ILSVRC Video dataset for our training and OTB and VOT datasets for testing as in \cite{SiamFC}. In the 2015 challenge organized by ILSVRC \cite{ILSVRC2015}, a new dataset is presented for the challenge, namely ``Object Detection from Video'', which has more than $4000$ videos. In each video, an object out of $30$ classes acts and the bounding box for each frame is provided. This rich amount of annotated data is utilized to generate our $200K$ triplet samples (the desired response, the localized and unlocalized patches) as it is done for VOT2015, and called as \textbf{CFCF ILSVRC} for short.

\subsection{CNN Architectures}
\subsubsection{The Custom Architectures for CFCF VOT2015 Dataset}
\begin{figure}
\begin{subfigure}{1\linewidth}
  \centering
  \includegraphics[width=1.0\linewidth]{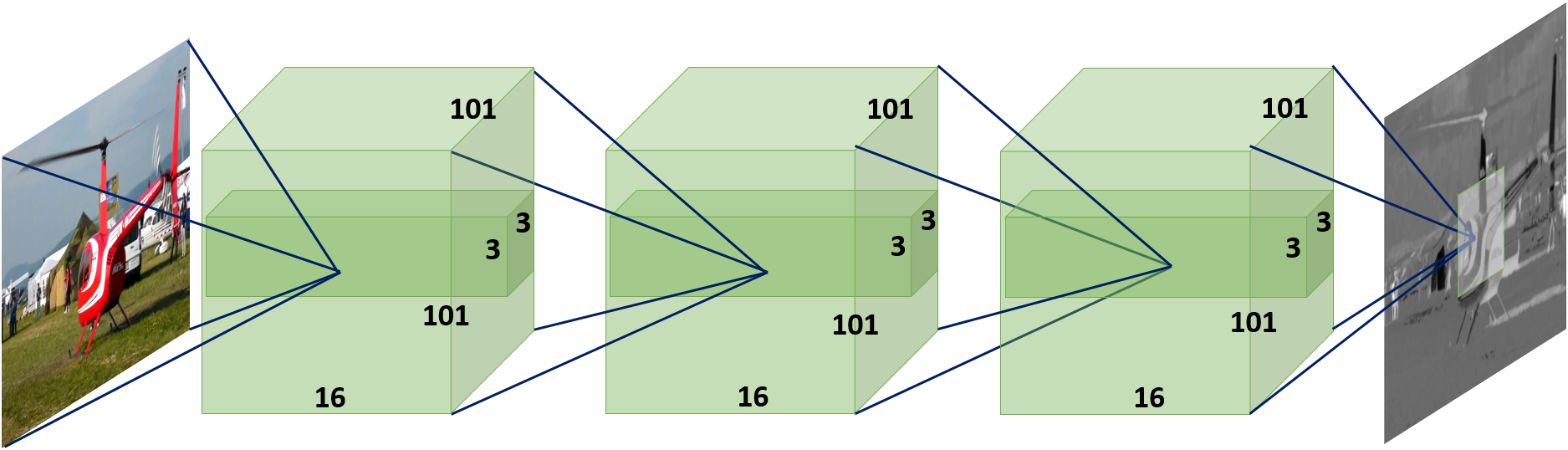}
  \caption{Single channel architecture}
  \label{Arch1}
\end{subfigure}%
\\
\begin{subfigure}{1.0\linewidth}
  \centering
  \includegraphics[width=1.0\linewidth]{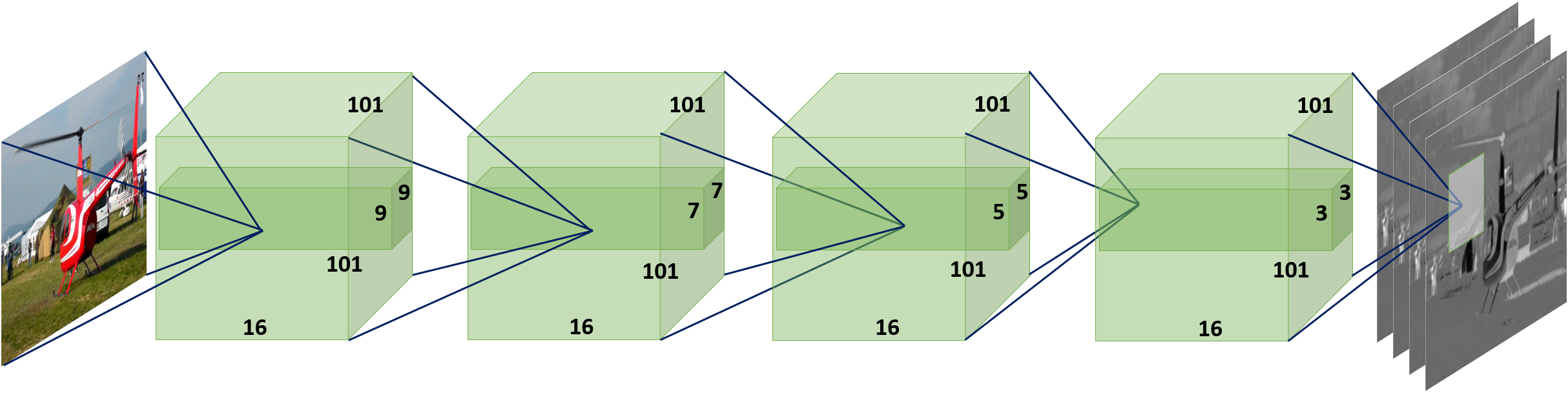}
  \caption{Multiple channel architecture}
  \label{Arch2}
\end{subfigure}
\caption{The custom architectures which are trained on the CFCF VOT2015.}
\label{ourarch}
\end{figure}

Two custom architectures are designed and illustrated in Figure \ref{Arch1} and \ref{Arch2}. The first one outputs a single feature map, whereas the second one yields multiple feature maps. The trackers utilizing these networks and the tracker DSST \cite{DSST} will be called \textbf{DSST\_CFCF} and \textbf{DSST\_MCFCF} for single and multiple channel correlation filters, respectively. CFCF VOT2015 is a medium scale dataset. Hence, we opt to design a relatively small architecture with respect to the state-of-the-art CNNs, such as \cite{VGG}. For this purpose, the input to our network is $3$ channel input image in $101\times 101$ dimensions. Our architecture consists of $4$ convolutional layers. All of these layers have a batch normalization layer after the convolutional layer part. The first three of them have a rectified linear unit (ReLU) \cite{ImageNetCNN} layer with a leak of $0.1$ \cite{prelu, prelu2}. In order to keep the spatial size of the feature maps constant, convolutional layers have the appropriate padding (\emph{e.g.} padding value is $1$ for a $3 \times 3$ kernel). The number of feature maps are shown in Figure \ref{ourarch}. The final layer outputs the map which will be utilized for the correlation task ($x(\theta)$ and $y(\theta)$ of Figure \ref{algflow}).
\subsubsection{Fine-tuning VGG-M \cite{VGG} for CFCF ILSVRC Dataset}
Unlike the network described above, the VGG-M network in \cite{VGG} is exploited in such a way that this network is cut from the first fully connected layer (\emph{fc6}), since the built framework only accepts the convolutional layers due to their shift invariance property. Although any other network model could be selected, VGG-M is fine-tuned to fairly compare against the CCOT tracker \cite{CCOT}, which utilizes the zeroth, the first and the fifth convolutional layers of VGG-M. In the literature, there exist efforts for the investigation of the useful feature maps. For instance, the study in \cite{HCF} performs analysis on the effect of different convolutional layers of AlexNet \cite{ImageNetCNN} and VGG-VD \cite{VGG_verydeep} models in a correlation filter based tracking framework, where conv-5 layer results in better performance than conv-3 and conv-4. Moreover, the analysis in deepSRDCF \cite{deepSRDCF} also verifies that the conv-5 layer of \cite{VGG} is the best performing layer. Since our baseline tracker CCOT is based on the tracker deepSRDCF \cite{deepSRDCF}, we prefer utilizing conv-1 and conv-5 layers along with the RGB channels of the image.

\begin{figure}[ht]
\centering
\includegraphics[width=0.48\textwidth]{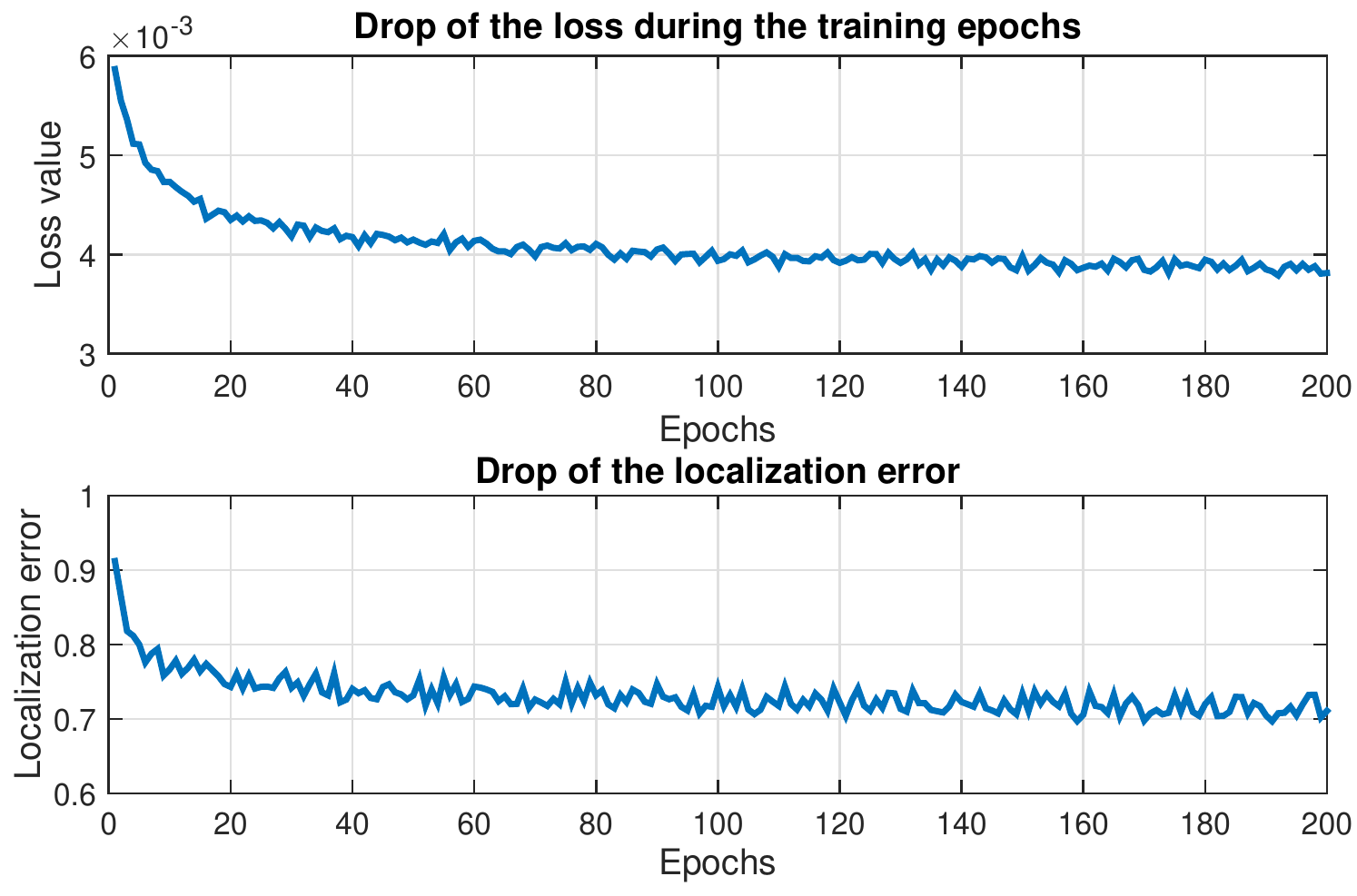}
\caption{The drop of the loss in \eqref{eqCost} and the localization error during the training epochs for fine-tuning VGG-M network.}
\label{costdrop}
\normalsize
\end{figure}

In order to train convolutional layers of VGG-M, an auxiliary layer with $32$ feature maps is added as the layer to be optimized by our cost function in \eqref{eqCost} by following the discussion in Section \ref{secComplexity} and Appendix \ref{Appendix2}. This augmentation is necessary, because the final convolutional layer of VGG-M has $512$ feature maps and the training with respect to the proposed loss becomes infeasible. The tracker obtained by integrating fully convolutional layers of VGG-M \cite{VGG}, fine-tuned in CFCF ILSVRC dataset with our cost function, into CCOT is simply called \textbf{CFCF}. The decrease of the cost in \eqref{eqCost} and the localization error, \emph{i.e.}, Euclidean distance between the desired and estimated object location, are plotted in Figure \ref{costdrop} during fine-tuning VGG-M network on CFCF ILSVRC dataset. During training, $51$\% of the time is spent for calculating the gradient terms of the loss function ($\frac{d\mathcal{L}}{dx^l}$ and $\frac{d\mathcal{L}}{dy^l}$), whereas $14$\% and $34$\% of the time are spent for forward and backward propagation of the convolutional parts of the model, respectively. The remaining time is consumed by other stages such as updating the weights ($3.5$\%).
\subsection{Evaluation in OTB-2013 by Training on CFCF VOT2015}
\subsubsection{Comparison with respect to hand-crafted features}
In order to understand the effects of different feature types on the tracking performance, a comparative analysis has been carried out. For this purpose, we work on DSST \cite{DSST}, which is a state-of-the-art multi-channel CFB tracker with a scale search support. For fair comparisons, the only revised part in this tracker is its feature extraction stage.
\begin{table*}
\caption{\label{meanOPDP}\small Analysis on feature type and quantity. \textbf{Raw}: raw image intensities, \textbf{X Grad}: magnitude of the horizontal gradient, \textbf{Y Grad}: magnitude of the vertical gradient, \textbf{CFCF}: learned single feature, \textbf{MCFCF}: learned multiple features. \textbf{Mean OP}: average overlap score with the threshold 0.5, \textbf{Mean DP}: average center location error with the threshold $20$ pixels.}
\normalsize
\adjustbox{max width=\linewidth}{
\renewcommand{\arraystretch}{1.0}
\begin{tabular}{c|c|c|c|c|c}
          & \textbf{DSST\_GRAY} & {\textbf{DSST} \textbf{\_GRAY\_GRADS}}& \textbf{DSST \cite{DSST}}& {\textbf{DSST\_CFCF}  (Proposed)} & {\textbf{DSST\_MCFCF}  (Proposed)}\\  \midrule \midrule
\rule{0pt}{12pt} \textbf{Feature types} & Raw & {Raw + X Grad  + Y Grad} & HOG + Raw & {Single CFCF + Raw + X Grad + Y Grad} & {MCFCF + Raw + X Grad + Y Grad} \vspace{0.4mm} \\  \hline
\textbf{\# of feature maps} & $1$ & $3$ & $28$ & $4$ & $11$\\ \hline
\textbf{Mean OP}\% & 62.2       & {\color[HTML]{009901} \textbf{67.4}} & {{67.3}} & {\color[HTML]{3531FF} \textbf{70.4}} & {\color[HTML]{FE0000} \textbf{70.9}} \\ \hline
\textbf{Mead DP}\% & 69.2       & {{70.0}} & {\color[HTML]{FE0000} \textbf{75.0}} & {\color[HTML]{3531FF} \textbf{74.8}} & {\color[HTML]{009901} \textbf{74.6}} \\ \hline
\end{tabular}}
\end{table*}
Tracking performance of different feature configurations are presented in Table \ref{meanOPDP}. The proposed single and multiple feature map configurations (DSST\_CFCF and DSST\_MCFCF, respectively) perform favorably against the use of hand-crafted features in terms of mean OP and mean DP, although the number of maps are fewer than the hand-crafted ones of DSST. In other words, DSST\_CFCF and DSST\_MCFCF have $4$ and $11$ feature maps, respectively, while DSST employs 28 HOG maps. DSST\_MCFCF has the best performance among the compared feature combinations.
\subsubsection{Comparison with respect to the state-of-the-art trackers}
For comparing our learned features against the recently proposed trackers, CCOT \cite{CCOT} (winner of VOT2016), which allows the use of multi-resolution feature maps, is adopted. For this purpose, we integrate our last layer features as well as the zeroth and first layers after the ReLU part, resulting in $27$ feature maps compared to $611$ feature maps of CCOT \cite{CCOT}. This configuration, called \textbf{MCFCF\_CCOT}, is also compared against deepSRDCF \cite{deepSRDCF} (the $2^{nd}$ best of VOT2015 challenge) utilizing $96$ feature maps of \cite{VGG}. A recent work SiamFC \cite{SiamFC}, where a fully convolutional model is trained for sliding window matching, is also compared with the proposed method.

\begin{figure}
\begin{minipage}[t]{0.495\linewidth}
\centering
\includegraphics[width=1\textwidth]{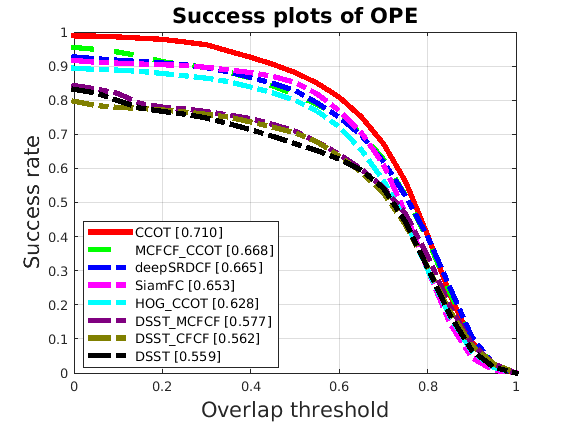} 
\end{minipage}
\hfill
\begin{minipage}[t]{0.495\linewidth}
\centering
\includegraphics[width=1\textwidth]{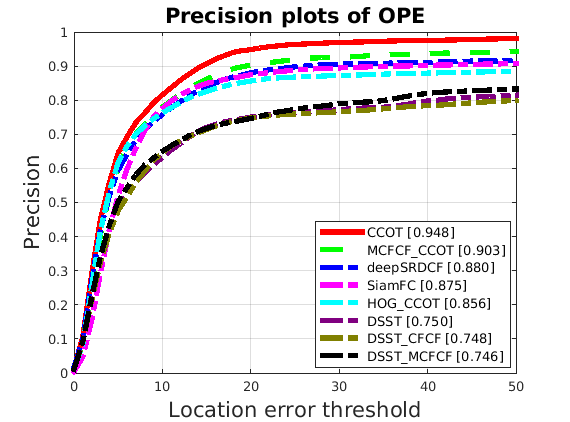} 
\end{minipage}
\caption{\label{OPEFIG}\small One-Pass-Evaluation (OPE) curves. \textbf{Left:} Overlap precision (OP) and \textbf{right}: center localization error (CLE).}
\end{figure}
\normalsize
Figure \ref{OPEFIG} presents OPE results on $40$ sequences of \cite{Benchmark2013}. Regarding average OP values (the left of Figure \ref{OPEFIG}), the proposed $27$ feature maps yield a close performance to CCOT with $611$ features. Meanwhile, it outperforms deepSRDCF, which utilizes $96$ feature maps. On the other hand, the proposed method performs favorably against deepSRDCF and SiamFC in terms of CLE values (the right of Figure \ref{OPEFIG}).

In Table \ref{AUCPerAttr}, AUC values of OP are presented for $11$ attributes. For most attributes, the proposed features perform close to CCOT, such as in the sequences with scale variation (SV), deformation (DEF) and background clutter (BC).
\begin{table}
\caption{\label{AUCPerAttr}\small AUC values for 11 attributes of the 40 sequences \cite{Benchmark2013}.}
\normalsize
\adjustbox{max width=\columnwidth}{
\renewcommand{\arraystretch}{1.5}
\begin{tabular}{c|c|c|c|c|c|c|c|c|c|c|c|c}
 & \textbf{IV} & \textbf{SV} & \textbf{OCC} & \textbf{DEF} & \textbf{MB} & \textbf{FM} & \textbf{IPR} & \textbf{OPR} & \textbf{OV} & \textbf{BC} & \textbf{LR} & \textbf{Avg.} \\ \hline
 MCFCF\_CCOT (\textbf{ours})& 0.63 & 0.66 & 0.63 &0.70&0.69&0.63&0.64&0.64&0.58&0.65&0.62&\underline{0.67}\\ \hline
 CCOT \cite{CCOT} & 0.70 &  0.70 & 0.70 & 0.70 &0.76&0.71&0.69&0.69&0.76&0.68&0.75&\textbf{0.71}				\\ \hline
 deepSRDCF \cite{deepSRDCF} & 0.62 & 0.64 &0.63 &0.68 &0.69&0.64&0.65&0.65&0.64&0.65&0.44&0.67			\\ \hline
 SiamFC \cite{SiamFC} & 0.57 &  0.61 &0.63&0.60&0.60&0.60&0.65&0.63&0.65&0.60&0.63&0.65			\\ \hline
 HOG\_CCOT & 0.55 & 0.58& 0.61 & 0.65 &0.64&0.57&0.59&0.59&0.53&0.61&0.53&0.63			\\ \hline
 MCFCF\_DSST (\textbf{ours})& 0.56 & 0.56 &0.55&0.58&0.53&0.50&0.57&0.55&0.47&0.55&0.56&0.58			\\ \hline
 CFCF\_DSST (\textbf{ours})&0.56 & 0.56 &0.54&0.55&0.50&0.47&0.56&0.54&0.50&0.52&0.50&0.56			\\ \hline
 DSST \cite{DSST} &0.56 & 0.56 &0.53&0.52&0.52&0.49&0.58&0.53&0.47&0.52&0.51&0.56			\\ \hline
\end{tabular}}
\end{table}
\subsection{Evaluations in OTB-2013 and OTB-2015 by Training on CFCF ILSVRC}
The fine-tuned VGG-M is tested on OTB-2013 with $51$ videos and OTB-2015 with $100$ videos. As in \cite{CCOT}, the zeroth, first and fifth convolutional layers of VGG are employed. For the remaining part of the simulations, CCOT \cite{CCOT} and SAMF \cite{SAMF} are utilized to integrate our learned features, and the proposed configurations are denoted as \textbf{SAMF\_CFCF} and \textbf{CCOT\_CFCF}, respectively. Moreover, we re-run the CCOT configuration that the authors use for their ECCV submission \cite{CCOT}, since the results might change on different CPUs. This configuration is named as \textbf{CCOT\_VGG}, while \textbf{SAMF\_VGG} is SAMF that employs VGG features as described earlier.

\begin{figure}
\centering
\includegraphics[width=0.9\linewidth]{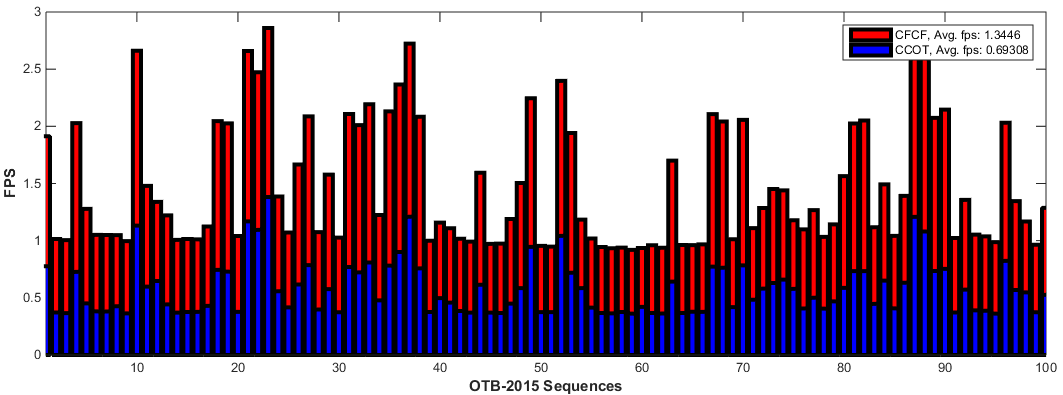}
\caption{Speed comparison between CCOT (baseline) and the proposed tracker CFCF on OTB-2015 sequences.}
\label{Speedy}
\end{figure}

For SAMF\_CFCF, the default parameters of \cite{SAMF} is used except for the learning rate, which is halved, since the utilized features (VGG and CFCF), which are more robust than hand-crafted ones, perform better with relatively lower learning rates. Similarly, for CCOT\_CFCF, the fine-tuned VGG-M network is integrated into the CCOT tracker by using its default hyper-parameters, except for the number of Conjugate Gradient iterations to compute the correlation filter. Hence, the best practice is to decrease the overfitting factor. The default iteration number of CCOT is $5$, whereas we only perform $1$ conjugate gradient iteration except for the first frame (which has $100$ iterations in both our case and the baseline CCOT configuration). Hence, the learned features also help to double the computation speed, as the fps values are reported in Figure \ref{Speedy} for $100$ videos of OTB-2015. The fps is measured in Intel Xeon E5 2623 3.0 GHz except for the CNN feature extraction part which is performed by NVidia Tesla K40 in MatConvNet \cite{MatConvNet}. For a $200 \times 200$ target object, the speed of the proposed tracker is around 1.7 fps. During tracking, $55\%$ of the time is consumed for correlation filter learning, $17\%$ and $11\%$ of the time pass during the CNN feature extraction and the detection of the object, respectively, while the remaining part is spent for other functions such as image resizing.
\begin{figure}
\begin{minipage}{1\linewidth}
\centering
\includegraphics[width=1\linewidth]{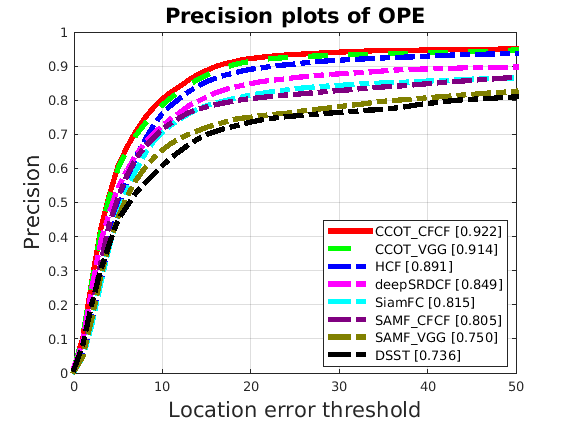} 
\caption{OTB-2013 localization error curves}
\label{OTB2013Prec}
\end{minipage}
\\
\begin{minipage}{1\linewidth}
\centering
\includegraphics[width=1\linewidth]{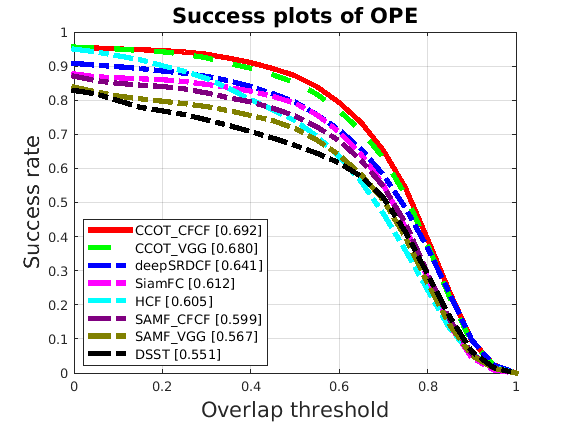} 
\caption{OTB-2013 overlap curves}
\label{OTB2013AUC}
\end{minipage}
\end{figure}
\normalsize
\begin{figure}
\begin{minipage}{1\linewidth}
\centering
\includegraphics[width=0.9\linewidth]{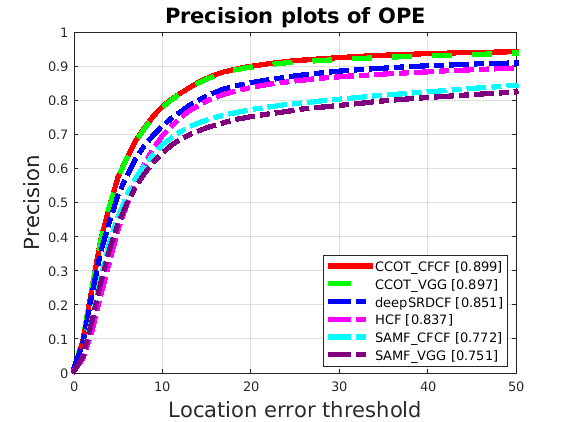} 
\caption{OTB-2015 localization error curves}
\label{OTB2015Prec}
\end{minipage}
\\
\begin{minipage}{1\linewidth}
\centering
\includegraphics[width=0.9\linewidth]{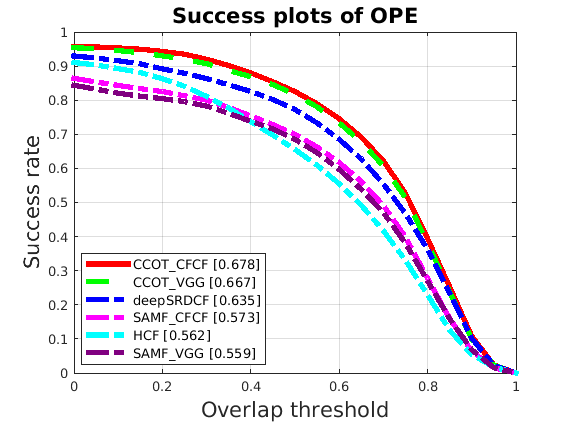} 
\caption{OTB-2015 overlap curves}
\label{OTB2015AUC}
\end{minipage}
\end{figure}

Figure \ref{OTB2013Prec} and \ref{OTB2013AUC} present the localization error and overlap curves for OTB-2013 dataset, respectively. Moreover, Figure \ref{OTB2015Prec} and \ref{OTB2015AUC} show the localization error and overlap curves for OTB-2015 dataset, respectively. In both of the datasets, the proposed features maintain its superiority over the baseline trackers SAMF and CCOT, while performing favourably or comparably against the state-of-the-art CFB trackers deepSRDCF \cite{deepSRDCF} and HCF \cite{HCF}.
\begin{table}
\caption{\label{AUCPerAttr2015}\small AUC values for 11 attributes for 100 sequences of \cite{BenchmarkPAMI}.}
\normalsize
\adjustbox{max width=\columnwidth}{
\renewcommand{\arraystretch}{1.5}
\begin{tabular}{|c|c|c|c|c|c|c|c|c|c|c|c|}
\hline
 & \textbf{IV} & \textbf{SV} & \textbf{OCC} & \textbf{DEF} & \textbf{MB} & \textbf{FM} & \textbf{IPR} & \textbf{OPR} & \textbf{OV} & \textbf{BC} & \textbf{LR} \\ \hline
CCOT\_CFCF (\textbf{ours})&\textbf{70.0}&\textbf{66.1}&\textbf{67.2}&\textbf{61.4}&\textbf{71.7}&\textbf{67.8}&\textbf{64.4}&\textbf{65.3}&\textbf{66.0}&65.2&\textbf{60.5}\\ \hline
CCOT\_VGG \cite{CCOT}&67.9&65.5&66.0&60.7&69.7&67.1&62.1&64.5&63.0&\textbf{66.0}&58.8\\\hline
SAMF\_CFCF \textbf{ours} &55.5&52.6&55.7&50.2&58.0&53.1&56.3&55.6&51.4&59.4&47.6\\\hline
SAMF\_VGG &56.8&51.1&54.5&49.6&56.7&53.5&54.0&54.5&55.1&{56.4}&44.4\\\hline
deepSRDCF \cite{deepSRDCF}&62.4&60.7&60.3&56.7&64.2&62.8&58.9&60.7&55.3&62.7&47.5 \\\hline
HCF \cite{HCF}&54.1&48.5&52.6&53.0&58.5&57.0&55.9&53.4&47.4&58.5&43.9\\\hline
\end{tabular}}
\end{table}
Per-attribute Area-Under-Curve (AUC) values are reported in Table \ref{AUCPerAttr2015}. Except for background clutter (BC), our proposed tracker performs favorably against CCOT. When the best performing configurations of the concurrent works \cite{CFNet} and \cite{DCFNet} are compared against our tracker on OTB2015, a significant amount of performance increase is observed. DCFNet \cite{DCFNet} has AUC of 57.5 and CFNet \cite{CFNet} has AUC of 58.9, whereas we achieve AUC of 67.8 out of 100.0. This is probably due to the fact that the concurrent works mainly focus on speed while we target at accuracy.
\subsection{Evaluation in VOT2016 by Training on CFCF ILSVRC}
As it is mentioned in the previous section, the proposed features that are integrated into CCOT has also been tested in VOT2016 challenge dataset including 60 videos. For making fair comparison between the fine-tuned VGG features for our loss function and the VGG features utilized by CCOT, VOT2016 challenge configuration of CCOT is utilized. In that configuration, first and fifth convolutional layers of VGG are employed as well as the color names of \cite{ColorNames} with $11$ features and $31$ HOG gradient maps in \cite{SRDCF}.
\begin{table}
\caption{VOT2016 performance results. Unlike OTB experiments using only convolutional layers, CCOT and our tracker CFCF employ convolutional features, color names, and HOG orientation maps (c.f. Section \ref{secAblation} for the performance comparison with only convolutional layers).}
\label{tableVOT2016}
\centering
\adjustbox{max width=\columnwidth}{
\renewcommand{\arraystretch}{1.5}
\begin{tabular}{|c|c|c|c|c|c|}
\hline
\textbf{Trackers} & \textbf{EAO} & \textbf{Acc. Rank} & \textbf{Rob. Rank}& \textbf{Acc. Raw} & \textbf{Fail. Raw}\\\hline \hline
\textbf{CFCF} & \first{\textbf{0.3903}} & 1.98 & 2.27 & \third{\textbf{0.54}} & \first{\textbf{0.63}} \\\hline
\textbf{CCOT} \cite{CCOT} & \second{0.3310} & 2.55 & 2.95 & 0.52 & \third{0.85} \\\hline
\textbf{TCNN} \cite{TCNN} & \third{0.3249} & 1.97 & 3.92 & 0.54 & 0.96 \\\hline
\textbf{SSAT} \cite{MDNET,VOT2016} & 0.3207 & 1.62 & 3.80 & \first{0.57} & 1.04\\\hline
\textbf{MLDF} \cite{VOT2016} & 0.3106 & 3.70 & 2.82 & 0.48 & \second{0.83} \\\hline
\textbf{Staple} \cite{STAPLE} & 0.2952 & 2.57 & 4.83 & 0.54 & 1.35\\\hline
\textbf{DDC} \cite{VOT2016} & 0.2929 & 2.27 & 4.62 & 0.53 & 1.23\\\hline
\textbf{EBT} \cite{EBT} & 0.2913 & 5.07 & 2.88 & 0.4 & 0.90\\\hline
\textbf{SRBT} \cite{VOT2016} & 0.2904 & 3.73 & 4.47 & 0.50 & 0.125\\\hline
\textbf{STAPLEp} \cite{VOT2016} & 0.2862 & 2.03 & 4.42 & \second{0.55} & 1.32\\\hline
\textbf{DNT} \cite{VOT2016} & 0.2783 & 3.03 & 4.47 & 0.50 & 1.18\\\hline
\end{tabular}}
\end{table}
Table \ref{tableVOT2016} reports the performance results of VOT2016 challenge. Among $71$ participants, we only show the top ten trackers and the proposed tracker CFCF ordered by the EAO metric unifying the robustness and accuracy of the trackers.

\begin{figure}[ht]
\centering
\includegraphics[width=0.4\textwidth]{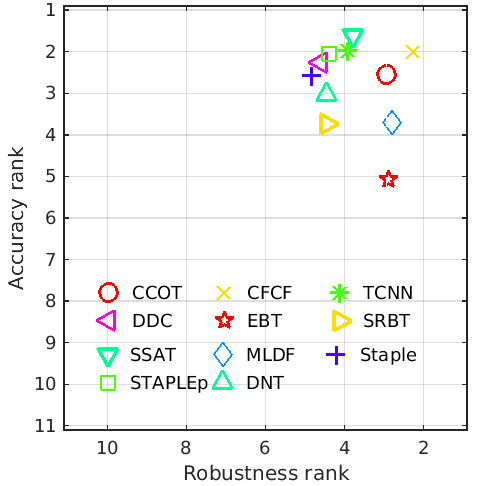}
\caption{Accuracy-Robustness Ranking plot. Closeness to the top right indicates good tracking performance.}
\label{ARPLOTVOT2016}
\end{figure}
In Figure \ref{ARPLOTVOT2016}, the ranking results in Table \ref{tableVOT2016} is pictured within a $2$-D plot. As the figure shows, the proposed tracker outperforms all of the existing participants. Moreover, the proposed features significantly improves the top tracker CCOT by $18.7$\% in terms of EAO. On the other hand, the number of optimization iterations is reduced to $1$ from $5$, bringing a significant decrease in the computation time. It should also be noted that the number of failures is decreased by $25$\% with respect to the CCOT. The raw accuracy performance is also improved by at least $3.5$\%. Table \ref{tableVOT2016Attr} reports the per-attribute accuracy and failure values of the compared trackers for VOT2016 dataset. The proposed tracker has favorable accuracy and failure values compared to CCOT for most of the attribute types. Notably, CFCF has significantly better accuracy than the compared trackers for camera motion attribute and less failures for scale change and camera motion attributes. Regarding the comparison against the concurrent works \cite{CFNet} and \cite{DCFNet}, knowing the fact that \cite{DCFNet} has an EAO value below 0.25 (in VOT2015) and VOT2016 annotations are more challenging, the possible EAO value of the tracker in \cite{DCFNet} would be smaller than our EAO value. Since CFNet \cite{CFNet} exploits VOT2016 dataset as the validation dataset, we do not compare CFNet \cite{CFNet} and our tracker on VOT2016 dataset.

\begin{table}
\caption{Per-attribute results of Accuracy/Failures on VOT2016. Bold indicates the best.}
\label{tableVOT2016Attr}
\centering
\adjustbox{max width=\columnwidth}{
\renewcommand{\arraystretch}{1.5}
\begin{tabular}{|c|c|c|c|c|c|}
\hline
& \textbf{cam. mot.} & \textbf{ill. ch.} & \textbf{mot. ch.}& \textbf{occ.} & \textbf{scal. ch.}\\\hline \hline
\textbf{CFCF} & \textbf{0.57}/\textbf{17.0} & 0.64/2.0&0.50/15.0&0.47/{15.0}&0.53/\textbf{6.0}\\\hline
\textbf{CCOT} \cite{CCOT}&0.56/24.0&0.65/2.0&0.47/20.0&0.44/\textbf{14.0}&0.50/13.0\\\hline
\textbf{TCNN} \cite{TCNN}&0.55/27.9&0.64/3.1&0.52/22.1&0.51/15.3&0.51/14.9\\\hline
\textbf{SSAT} \cite{MDNET,VOT2016}&0.57/34.1&0.67/2.3&\textbf{0.54}/21.7&\textbf{0.51}/23.7&\textbf{0.55}/15.1\\\hline
\textbf{MLDF} \cite{VOT2016}&0.51/22.0&0.58/2.0&0.45/19.0&0.41/17.0&0.44/7.0\\\hline
\textbf{Staple} \cite{STAPLE}&0.55/34&\textbf{0.71}/7.0&0.51/35.0&0.43/24.0&0.51/15.0\\\hline
\textbf{DDC} \cite{VOT2016}&0.56/27.0&0.59/5.0&0.52/24.0&0.45/18.0&0.49/14.0 \\\hline
\textbf{EBT} \cite{EBT} &0.49/20.0&0.41/3.0&0.44/19.0&0.37/17.0&0.36/11.0 \\\hline
\textbf{SRBT} \cite{VOT2016}&0.49/33.0&0.44/\textbf{1.0}&0.46/24.0&0.43/20.0&0.43/13.0 \\\hline
\textbf{STAPLEp} \cite{VOT2016}&0.56/35.0&0.69/6.0&0.51/33.0&0.44/23.0&0.53/17.0\\\hline
\textbf{DNT} \cite{VOT2016}&0.52/31.8&0.53/2.0&0.49/21.1&0.44/20.1&0.48/11.6\\\hline
\end{tabular}}
\end{table}

\subsection{Ablation Study}
\label{secAblation}
We conduct a set of ablation studies on VOT2016 dataset to analyze how the proposed feature learning framework perform in different training and input configurations. During these experiments, only convolutional features are employed. 

\textbf{Utilization of chroma components:} Table \ref{tableAblation1} compares the performances of our tracker configurations for the inputs with color and gray scale when the zeroth, first and fifth convolutional layer activations are utilized. Moreover, it also demonstrates the improvement obtained by fine-tuning the VGG-M network with our loss function. When the network is fine-tuned by our framework, in terms of EAO, more than $11$\% improvement is achieved over the use of pre-trained network while increasing accuracy by $2$\% and decreasing the number of failures by $25$\%. If the input is grayscale image, the tracking performance decreases, however, it achieves a close EAO ($0.26$) to the top ten trackers among VOT2016 challenge participants. Moreover, fine-tuning VGG-M network improves EAO value by $12$\% over the one trained on ImageNet \cite{ImageNet}.
\begin{table}
\centering
\caption{The effect of fine-tuning VGG-M \cite{VGG} on the tracking performance (VOT2016) for grayscale and color inputs. Pre-trained means VGG-M trained on ImageNet \cite{ImageNet} dataset for classification task, and fine-tuned means Fine-tuning the pre-trained VGG-M on CFCF ILSVRC dataset for the proposed correlation filter based tracking loss function.}
\label{tableAblation1}
\adjustbox{max width=\columnwidth}{
\renewcommand{\arraystretch}{1.5}
\begin{tabular}{|l|l|l|l|l|}
\hline
\textbf{Feature Type}         & \textbf{Input Type} & \textbf{Acc.} & \textbf{Fail.} & \textbf{EAO} \\ \hline
{Fine-tuned}  & RGB                 & \first{0.53}          & \first{0.75}           & \first{0.3398}        \\ \hline
{Pre-trained} & RGB                 & \third{0.52}          & \second{1.00}           & \second{0.3050}        \\ \hline
{Fine-tuned}  & Grayscale          & \second{0.52         } & \third{1.32}           & \third{0.2638}        \\ \hline
{Pre-trained} & Grayscale          & {0.51} & 1.47           & 0.2354        \\ \hline
\end{tabular}}
\end{table}

\textbf{Performance analyses of different layers:} In this part, the tracking performances of different layers of our fine-tuned network model (VGG-M) are investigated. For this purpose, convolutional layers are individually tested on VOT2016 dataset in terms of accuracy, robustness (failures) and EAO. Then, the best performing higher level (among (conv-3, conv-4 and conv5) convolutional layer and lower level convolutional layer (among conv-1 and conv-2) are merged to obtain a boosted performance.
\begin{table}
\centering
\caption{Comparison between different convolutional layers of the learned network (VGG-M) by the proposed loss function on VOT2016.}
\label{tableAblation3}
\adjustbox{max width=\columnwidth}{
\renewcommand{\arraystretch}{1.5}
\begin{tabular}{|c|c|c|c|}
\hline
\textbf{\# of samples} & \textbf{Accuracy} & \textbf{Failures} & \textbf{EAO} \\ \hline
\textit{conv-1} &{0.52} &0.97 &0.3062 \\ \hline
\textit{conv-2} &\second{0.53} &\third{0.90} &\third{0.3187} \\ \hline
\textit{conv-3} &{0.50} &{1.10} &{0.2719} \\ \hline
\textit{conv-4} &{0.50} &{0.98} &{0.2895} \\ \hline
\textit{conv-5} &{0.50} &{1.02} &{0.2994} \\ \hline
\textit{conv-2+conv-5} &\first{0.54} &\first{0.82} &\first{0.3364} \\ \hline
\textit{conv-1+conv-5} &\third{0.53} &\second{0.87} &\second{0.3269} \\ \hline
\end{tabular}}
\end{table}
Table \ref{tableAblation3} compares EAO, accuracy and robustness values of different layers and some combinations. \textit{Conv-2} layer has the best EAO, accuracy and robustness values. Among the higher convolutional layers, \textit{conv-5} performs better than \textit{conv-4}. Hence, the combination of \textit{conv-2} and \textit{conv-5} performs favorably against \textit{conv-2} or \textit{conv-5}.

\textbf{Training from scratch:} We analyze the impact of the size of the training set on the tracking performance. For this purpose, the first four convolutional layers of VGG-M are utilized. The final and the fifth layer of the model is the convolutional layer with $32$ feature maps. For the tracking application, these $32$ feature maps and color channels of the input image are employed in CCOT implementation.
\begin{table}
\centering
\caption{The effect of training the first four convolutional layers of VGG-M \cite{VGG} with the augmented convolutional layer with $32$ feature maps from scratch on the tracking performance (VOT2016) is reported for varying amount of training data (CFCF ILSVRC).}
\label{tableAblation2}
\adjustbox{max width=\columnwidth}{
\renewcommand{\arraystretch}{1.5}
\begin{tabular}{|l|l|l|l|}
\hline
\textbf{\# of samples} & \textbf{Accuracy} & \textbf{Failures} & \textbf{EAO} \\ \hline
\textbf{None} &{0.47} &2.38 &0.1707 \\ \hline
\textbf{0.5K} &\third{0.47} &2.45 &0.1667 \\ \hline
\textbf{5K} &\first{0.48} &\third{2.32} &\third{0.1749} \\ \hline
\textbf{50K} &\second{0.47} &\second{2.22} &\second{0.1780} \\ \hline
\textbf{200K} &{0.47} &\first{2.05} &\first{0.1883} \\ \hline
\end{tabular}}
\end{table}
In Table \ref{tableAblation2}, the tracking performance on VOT2016 dataset improves in terms of average number of failures as the number of training samples are increased from $500$ to $200$K. The randomly initialized network performs favorably against the one trained with $500$ samples. One reason for the performance degradation by training a very small amount of data is overfitting. Remarkably, training $200$K samples significantly improves the EAO value and average failures over the random network by $10$\%, and these values are comparable against the participating trackers of VOT2016 which have average ranking even though the tracker configuration adopts 35 feature maps ($32$ from conv-5 and $3$ from color channels).

\begin{figure}
\begin{subfigure}{.48\linewidth}
  \centering
  \includegraphics[width=0.98\linewidth]{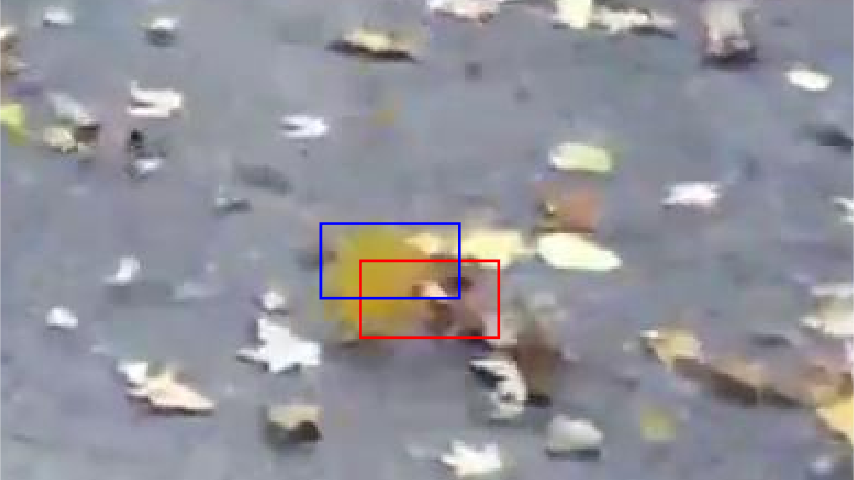}
  \caption{The sequence \emph{leaves}}
  \label{Qual1}
\end{subfigure}%
\begin{subfigure}{.48\linewidth}
  \centering
  \includegraphics[width=0.98\linewidth]{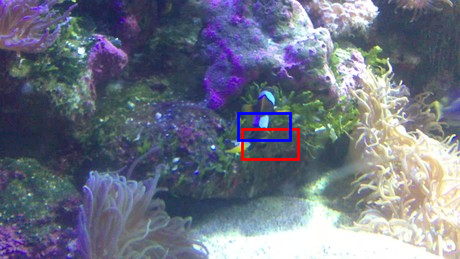}
  \caption{The sequence \emph{fish4}}
  \label{Qual2}
\end{subfigure}
\\
\begin{subfigure}{.48\linewidth}
  \centering
  \includegraphics[width=0.98\linewidth]{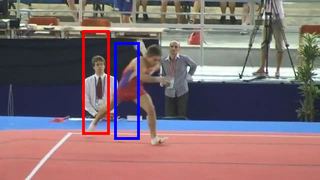}
  \caption{The sequence \emph{gymnastics1}}
  \label{Qual3}
\end{subfigure}%
\begin{subfigure}{.48\linewidth}
  \centering
  \includegraphics[width=0.98\linewidth]{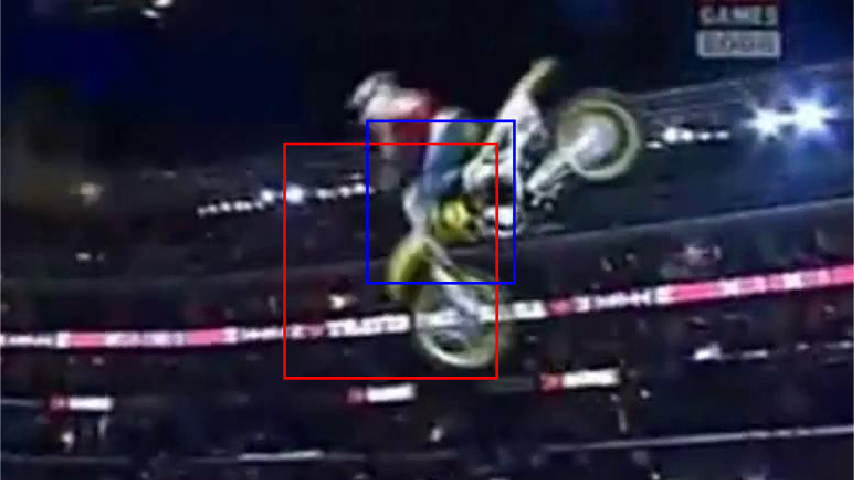}
  \caption{The sequence \emph{motocross2}}
  \label{Qual4}
\end{subfigure}
\caption{Failure cases of our approach in VOT2016 dataset. \first{Red} and \second{blue:} indicate CFCF (ours) and CCOT, respectively.}
\label{failureCases}
\end{figure}

\textbf{Failure cases:} Figure \ref{failureCases} illustrates some failure cases of the proposed tracker CFCF with respect to CCOT \cite{CCOT} when both trackers utilize convolutional, HOG gradient maps and color names. The sequences are selected according to the case when both of them fail at the same frame or do not fail.
Some failure cases show up in particular sequences though the proposed features improve the tracking performance over the ones extracted from the pre-trained model. Some possible reasons might be the amount of difference between challenges of VOT2016 and the training set CFCF ILSVRC (the sequence \emph{leaves}), abrupt aspect ratio change (the sequence \emph{motocross2} and \emph{gymnastics1}), and object deformation (the sequence \emph{fish4}).

\section{Conclusion and Future Work}
\label{secConclusion}
In this study, we address the feature learning problem for correlation filter based visual tracking task. For this purpose, a novel and generic framework is proposed to train any deep and fully convolutional network. By exploiting the correlation theorem, an efficient backpropagation formulation is presented to train any fully convolutional network by using stochastic gradient descent algorithm. The introduced feature learning method is trained on the frames generated by utilizing VOT2015 and ILSVRC Video datasets. The learned models have been integrated into the state-of-the-art correlation filter based trackers to show the validity of the proposed technique. In benchmark tracking datasets, favorable performance is achieved against the state-of-the-art tracking methods. Notably, the top performing tracker of VOT2016 challenge has been improved by at least $18$\% in terms of the expected average overlap metric. The proposed methodology can be adopted to custom deep network designs.
\appendices
\section{Derivation of $\frac{d\mathcal{L}}{dh^k}$ and $\frac{d\mathcal{L}}{dx^l}$}
\label{secAppendix1}
By the Correlation Theorem in \eqref{eqCorrTheorem}, $\mathcal{L}$ is:
\begin{equation}
\label{eqApp1}
\mathcal{L}=\sum\limits_n \left( \sum\limits_l \sum\limits_i  h^l[i]x^l[i+n]-g[n] \right)^2
\end{equation}
The partial derivative for a particular component of $h^k[m]$ is:
\footnotesize
\begin{equation}
\label{eqApp2}
\frac{\partial \mathcal{L}}{\partial h^k[m]} = \sum\limits_n \left(\sum\limits_l \sum\limits_i h^l[i]x^l[i+n]-g[n]\right) \frac{\partial \sum_l \sum_i h^l[i] x^l[i+n]}{\partial h^k[m]}
\end{equation}
\begin{equation}
\label{eqApp3}
\frac{\partial \mathcal{L}}{\partial h^k[m]} = \sum\limits_n \left(\sum\limits_l \sum\limits_i h^l[i]x^l[i+n]-g[n]\right) x^k[m+n]
\end{equation}
\normalsize
If we utilize the error signal $e[n]=\sum\limits_l \sum\limits_i h^l[i]x^l[i+n]-g[n]$ defined in \eqref{eqApp4}, the derivatives will have better interpretation for the sake of both the time and frequency domain. By substituting this error to \eqref{eqApp3}, the derivative signal will have an efficient calculation as follows by using \eqref{eqCorrTheorem}:
\begin{equation}
\label{eqApp5}
\frac{\partial \mathcal{L}}{\partial h^k[m]}=
\sum\limits_n e[n]x^k[m+n]=[\mathcal{F}^{-1}\{E^* \odot X^k\}][m]
\end{equation}
By utilizing the equation \eqref{eqCorrTheorem}, $\frac{\partial \mathcal{L}}{\partial x^l[m]}$ can be derived as follows:
\begin{equation}
\label{eqApp6}
\frac{\partial \mathcal{L}}{\partial x^l[m]}=
\sum\limits_n e[n]h^l[m-n]=[\mathcal{F}^{-1}\{E \odot H^l\}][m]
\end{equation}

\section{The effect of a layer on the correlation quality of the previous one}
\label{Appendix2}
In this part, it is analyzed that the correlation quality of a layer behaves analogous to the layer above it if some assumptions on the additive appearance noise hold. This noise can be perceived as the appearance difference between the template $x$ and the test patch $z$. Convolutional layers have a set of 2-D feature maps. To obtain another convolutional layer on top of the previous one, they are summed with a set of weight parameters. For this purpose, $X$ is 2-D DFT of a single feature map obtained from a network in a certain layer, \emph{e.g.} $l^{th}$ layer, for the training example $x$. Similarly, $z$ is the test patch with the centered object and the corresponding correlation filter for $x$ is
\begin{equation}
\label{App1}
H = \frac{X \odot \hat{G^*}}{X^* \odot X+\gamma},
\end{equation}
where $\gamma$ is the regularization parameter, and $\hat{G}$ is the DFT of the desired response $\hat{g}$ for the template $x$ with a peak in its center location. If the localized test sample $z$ has the feature map in DFT domain as $Z = X+\mu$ with $\mu$ being the additive noise due to the appearance change of the object, then the resulting correlation error turns out to be:
\begin{equation}
\label{App2}
\begin{split}
\mathcal{E}_{single} = H^* \odot Z - \hat{G} \quad \quad \quad\quad  \\
=\left( \frac{X^* \odot \hat{G}}{X^*\odot X+\gamma}\odot (X+\mu)- \hat{G} \right) \approx \frac{\mu X^* \odot \hat{G}}{X^* \odot X + \gamma}
\end{split}
\end{equation}

If the convolutional kernel at level $l-1$ is assumed to be $1 \times 1$ with their values fixed to $1$ and there exists only two feature maps, then we can split $X$ as $X=X_1+X_2$ by ignoring the bias terms. In this case, the feature map of the test example $Z$ will be split as $Z = Z_1 + Z_2$, where $Z_1=X_1+\mu_1$ and $Z_2=X_2+\mu_2$. The $\mu_1$ and $\mu_2$ are the individual additive noises of the feature maps. The two correlation filters are:
\begin{equation}
\label{App3}
\begin{split}
H_1 = \frac{X_1 \odot \hat{G^*}}{X_1^* \odot X_1 + X_2^* \odot X_2+ \gamma} \\ H_2 = \frac{X_2 \odot \hat{G^*}}{X_1^* \odot X_1 + X_2^* \odot X_2+ \gamma}
\end{split}
\end{equation}
The correlation of $z$ and $h$ yields:
\begin{equation}
\label{App4}
\begin{split}
\mathcal{E}_{multi}= H_1^*Z_1 + H_2^*Z_2 - \hat{G} \quad \quad \quad \quad \\
= \frac{X_1^* \odot \hat{G}}{X_1^* \odot X_1 + X_2^* \odot X_2+ \gamma} \odot (X_1 + \mu_1) + \\ \frac{X_2^* \odot \hat{G}}{X_1^* \odot X_1 + X_2^* \odot X_2+ \gamma} \odot (X_2 + \mu_2)-\hat{G}
\end{split}
\end{equation}
By neglecting the effect of $\gamma$ value, the error is reduced to
\begin{equation}
\label{App5}
\mathcal{E}_{multi}=\frac{\mu_1 X_1^* \odot \hat{G} + \mu_2 X_2^* \odot \hat{G}}{X_1^* \odot X_1 + X_2^* \odot X_2+ \gamma}
\end{equation}
To make a similarity between the \eqref{App2} and \eqref{App5}, the $X$ is replaced with $X_1+X_2$ and $\mu = \mu_1 + \mu_2$ in \eqref{App2}. Moreover, all of the terms are copied in \eqref{App5}. Finally, we obtain the following error for single and multiple channels:
\footnotesize
\begin{equation}
\label{App6}
\begin{split}
\mathcal{E}_{multi}=\frac{\mu_1 X_1^* \odot \hat{G} + \mu_2 X_2^* \odot \hat{G} + \mu_1 X_1^* \odot \hat{G} + \mu_2 X_2^* \odot \hat{G}}{X_1^* \odot X_1 + X_2^* \odot X_2+ \gamma + X_1^* \odot X_1 + X_2^* \odot X_2+ \gamma} \\
\mathcal{E}_{single}=\frac{\mu_1 X_1^* \odot \hat{G} + \mu_2 X_2^* \odot \hat{G} + \mu_1 X_2^* \odot \hat{G} + \mu_2 X_1^* \odot \hat{G}}{X_1^* \odot X_1 + X_2^* \odot X_2 + X_1^* \odot X_2 + X_2^* \odot X_1+ \gamma}
\end{split}
\end{equation}
\normalsize
In \eqref{App5} and \eqref{App6}, both errors are proportional to $\mu_1$ and $\mu_2$. Hence, if the sum of these two variables (\emph{i.e.} $\mu$) decreases, $\mu_1$ and $\mu_2$ have also tendency to decrease. This derivation can be extended to more than two feature maps, where the same assumption could hold. If the two correlation response errors in two consecutive layers are almost the same, one can argue that the mitigation of the appearance noise in one of the layers is likely to reduce the correlation response error in another one. Hence, training a fully convolutional model to reduce its correlation error with respect to the top layer will eventually increase the correlation quality in the lower layers. The experimental results have clearly shown that this analysis is practically valid for most scenarios.

\ifCLASSOPTIONcaptionsoff
  \newpage
\fi

\bibliographystyle{IEEEtran}
\bibliography{IEEEabrv,IEEEexample}

\clearpage
\begin{IEEEbiography}[{\includegraphics[width=1in,height=1in,clip,keepaspectratio]{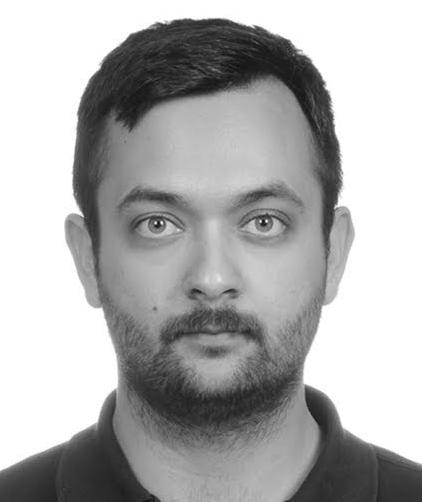}}]{Erhan Gundogdu}
received the B.Sc., M.Sc. and Ph.D. degrees in Electrical and Electronics Engineering Dept. from Middle East Technical University (METU), Ankara, Turkey, in 2010, 2012 and 2017, respectively. He was a Research Assistant at Electrical and Electronics Eng. Dept. at METU, Ankara, from 2010 to 2012, and is currently a Research Engineer in Aselsan Research Center, Ankara. His current research interests include visual tracking, object recognition, and detection.
\end{IEEEbiography}
\begin{IEEEbiography}[{\includegraphics[width=1.2in,height=1.2in,clip,keepaspectratio]{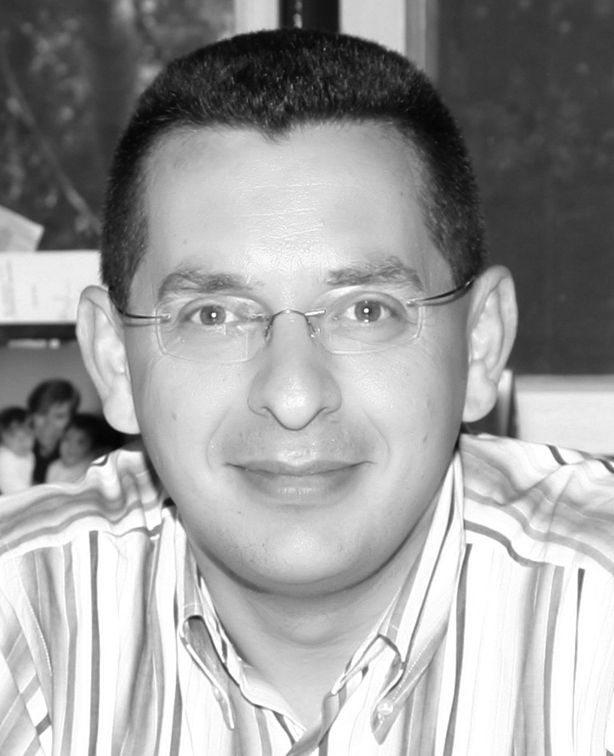}}]{A. Ayd\i n Alatan}
(SM’12) was born in Ankara, Turkey, in 1968. He received the B.Sc. degree from Middle East Technical University, Ankara, in 1990, the M.Sc. and D.I.C. degrees from the Imperial College of Science, Medicine and Technology, London, U.K., in 1992, and the Ph.D. degree from Bilkent University, Ankara, in 1997, in electrical engineering. He was a Post-Doctoral Research Associate with the Center for Image Processing Research, Rensselaer Polytechnic Institute, Troy, NY, USA, from 1997 to 1998, and the New Jersey Center for Multimedia Research, New Jersey Institute of Technology, Newark, NJ, USA, from 1998 to 2000. In 2000, he joined the faculty of the Department of Electrical and Electronics Engineering at Middle East Technical University. His current research interests include content-based video analysis, data hiding, robust image/video transmission over mobile channels and packet networks, image/video compression, video segmentation, 3D scene analysis, object recognition, detection and tracking.
\end{IEEEbiography}

\end{document}